%% file: CameraReady2027.tex
\documentclass[letterpaper]{article} 
\usepackage[preprint]{aaai2027}      
\usepackage{amsmath}
\usepackage{amssymb}
\usepackage{multirow}
\usepackage{xcolor}
\usepackage[hyphens]{url} 
\usepackage{graphicx}     
\usepackage{natbib}       
\usepackage{caption}      
\usepackage{booktabs}
\usepackage{algorithm}
\usepackage{algorithmic}

\usepackage{newfloat}
\usepackage{listings}

\DeclareCaptionStyle{ruled}{
  labelfont=normalfont,
  labelsep=colon,
  strut=off
} 

\floatstyle{ruled}
\newfloat{listing}{tb}{lst}{}
\floatname{listing}{Listing}
\definecolor{gainGreen}{HTML}{1A7F37}
\definecolor{lossRed}{HTML}{B3261E}

\definecolor{ruleGray}{HTML}{9A9A9A}
\definecolor{inkGray}{HTML}{555555}
\newcommand{\subwd}{\phantom{$_{\uparrow00.0}$}}

\newcommand{\gainsub}[1]{%
  \textcolor{gainGreen}{\rlap{$_{\uparrow#1}$}}\subwd
}

\newcommand{\dropsub}[1]{%
  \textcolor{lossRed}{\rlap{$_{\downarrow#1}$}}\subwd
}

\newcommand{\nosub}{\subwd}

\newcommand{\gainval}[1]{%
  \textcolor{gainGreen}{$\uparrow$#1}%
}

\newcommand{\dropval}[1]{%
  \textcolor{lossRed}{$\downarrow$#1}%
}
\newcommand{\bsec}[1]{\S\ref{#1}}
\newcommand{\btab}[1]{Table~\ref{#1}}
\newcommand{\bfig}[1]{Figure~\ref{#1}}
\newcommand{\beq}[1]{Eq.~\eqref{#1}}

\newcommand{\app}[1]{\S\ref{#1}}

\newenvironment{apptab}{%
  \par
  \addvspace{0.75\baselineskip}%
  \begingroup
  \centering
  \footnotesize
  \setlength{\tabcolsep}{4pt}%
  \renewcommand{\arraystretch}{1.12}%
}{%
  \par
  \endgroup
  \addvspace{0.10\baselineskip}%
}
\newenvironment{apptabsm}{%
  \par
  \addvspace{0.75\baselineskip}%
  \begingroup
  \centering
  \scriptsize
  \setlength{\tabcolsep}{3.5pt}%
  \renewcommand{\arraystretch}{1.15}%
}{%
  \par
  \endgroup
  \addvspace{0.10\baselineskip}%
}
\newcommand{\apptabcap}[2]{%
  \par
  \captionof{table}{#1}%
  \label{#2}%
  \par
  \addvspace{0.60\baselineskip}%
}

\newcommand{\appthinrule}{%
  \noindent
  \textcolor{ruleGray}{\rule{\linewidth}{0.4pt}}%
  \par
}

\newenvironment{promptcard}[2]{%
  \par
  \addvspace{0.60\baselineskip}%
  \appthinrule
  \addvspace{0.16\baselineskip}%
  {%
    \parfillskip=0pt
    \noindent
    {\footnotesize\bfseries #1}%
    \hfill
    {\scriptsize\color{inkGray}#2}%
    \par
  }%
  \addvspace{0.18\baselineskip}%
  \begingroup
  \scriptsize
}{%
  \endgroup
  \par
  \addvspace{0.10\baselineskip}%
  \appthinrule
  \addvspace{0.55\baselineskip}%
}
\newenvironment{priorquote}{%
  \begingroup
  \small
  \itshape
  \leftskip=1em
  \rightskip=0.4em
  \parindent=0pt
  \par
  \addvspace{0.15\baselineskip}%
}{%
  \par
  \addvspace{0.15\baselineskip}%
  \endgroup
}

\newcommand{\hdr}[2]{%
  \begin{tabular}[b]{@{}c@{}}
    #1\\
    {\scriptsize #2}
  \end{tabular}%
}
\newcommand{\ab}{\allowbreak}
\newcommand{\idtus}{\textunderscore\allowbreak}

\newcommand{\idt}[1]{%
  {\ttfamily\let\_\idtus #1}%
}
\newcommand{\naval}{%
  \textcolor{inkGray}{---}%
}
\title{SkillGLoW: Procedural-Family Skill Consolidation for Self-Improving Agents on Long-Horizon Task Streams}

\author{
Ao Yan\textsuperscript{\rm 1},
Zhang Xin\textsuperscript{\rm 2},
Jiawei Du\textsuperscript{\rm 2},
Joey Tianyi Zhou\textsuperscript{\rm 2}\corresponding
}
\affiliations{
\textsuperscript{\rm 1}National University of Singapore, Singapore\\
\textsuperscript{\rm 2}Institute of Advanced Intelligence and Computing (IAIC), Singapore
}

\begin{document}

\maketitle

\begin{abstract}
LLM agents increasingly self-improve by writing and reusing textual skills, kept either as one global document or as a flat pool of per-task entries, though most of the evidence comes from domains with structurally similar tasks. On long-horizon workloads where each task demands a different solution, the two forms fail in opposite ways: the document collapses into generic discipline, while the pool inflates and its entries stay bound to the instance that wrote them. We argue the missing unit of reuse is the solving procedure shared by a cluster of related tasks, and build SkillGLoW (Global--Local Weave) around it: the local skills a task writes from its own execution are aggregated into procedural families and compressed into de-instantiated global priors, while the instance detail they hold is regenerated per task rather than stored; a commit gate admits a prior only when real execution shows it does not degrade the deployed library. Across four benchmarks (mathematical reasoning, terminal automation, software repair, and embodied control) and three models, the priors gain 17.2 points (hard) over the no-skill baseline on average, with positive gains in all 12 continual-improvement runs, and 18.0 with local regeneration, while the library holds one prior per procedural family, $3.6\times$ more compact than the per-task pool. Under the same protocol GLoW leads a published single-document optimizer on 15 of 21 cells. Unmodified, the library lifts success on unseen ALFWorld tasks from 73.9\% to 83.9\%, evidence that what transfers is procedure rather than task memory.
\end{abstract}


\section{Introduction}

\begin{figure}[!t]
\centering
\includegraphics[width=\columnwidth]{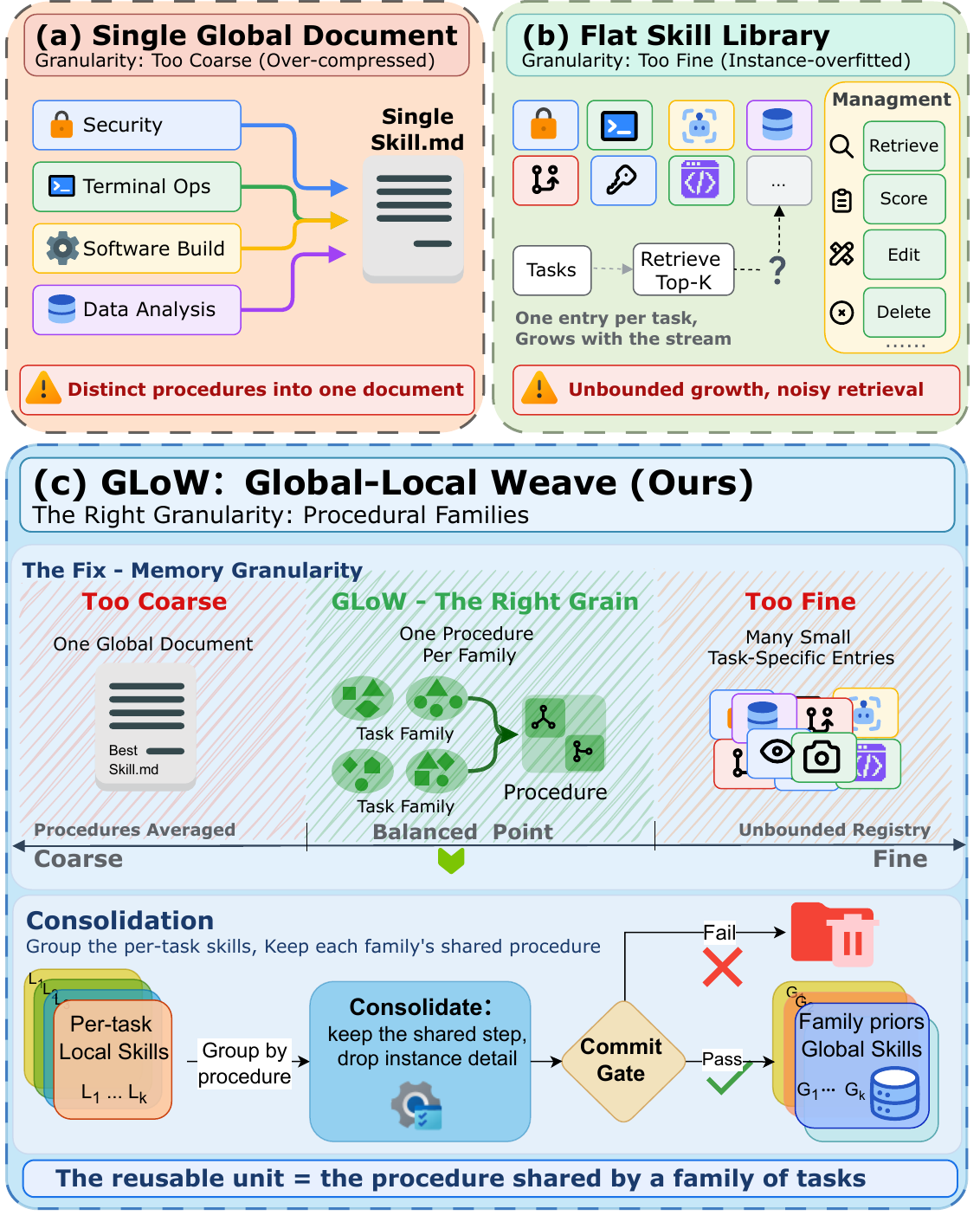}
\caption{The failure modes of single-document and flat-library skill organizations on heterogeneous long-horizon workloads, and the procedural-family unit that GLoW consolidates between them.}
\label{fig:intro}
\end{figure}

As base models and agent harnesses mature, LLM agents are moving from short-horizon, closed tasks toward longer-horizon, more complex environments
\citep{react,reflexion,agentbench,gaia,jimenez2024swebench,merrill2026terminalbench}.
Such tasks quickly outgrow fixed prompts and human-written skills, so recent systems distill reusable skills from their own execution trajectories \citep{mementoskills,museautoskill,trace2skill} and write successful or failed experience into documents reused by prompt injection \citep{acon,skillopt,memento}. Once skills are continually generated, revised, and reused, a more fundamental question arises: \emph{in what form should an agent organize and maintain these skills?}

SkillsBench offers direct evidence. Human-curated skills raise the pass rate by $16.2$ points, while skills a model writes for itself from the task description alone land $1.3$ points \emph{below} the no-skill baseline \citep{skillsbench}. Recent experience-driven methods therefore ground skills in real execution trajectories, but their evidence concentrates on structurally similar tasks; on long-horizon workloads, little beyond meta-level experience transfers across tasks \citep{memorytransfer}.

Every organizational form makes an implicit assumption about the task distribution. A single global document assumes that most tasks share one dominant procedure; a per-task skill library assumes that old entries can be reused wholesale. Both assumptions roughly hold in structurally similar domains, but they fail simultaneously on long-horizon workloads with heterogeneous solutions (Figure~\ref{fig:intro}): no dominant procedure exists, and few old entries fit new tasks \citep{memorytransfer,reme}, a pattern that survives when the pool is queried by retrieval (\S\ref{sec:ablation}). Both failures point to a missing unit of reuse, one that sits between the single document and the single entry. Solutions differ from instance to instance, but tasks are not isolated. Similar tasks form families and share a solving procedure, while each instance carries local constraints that surface only during execution. The shared procedure transfers across tasks; the instance details do not. An organizational scheme should keep both.

\begin{figure*}[!t]
\centering
\includegraphics[width=0.95\textwidth]{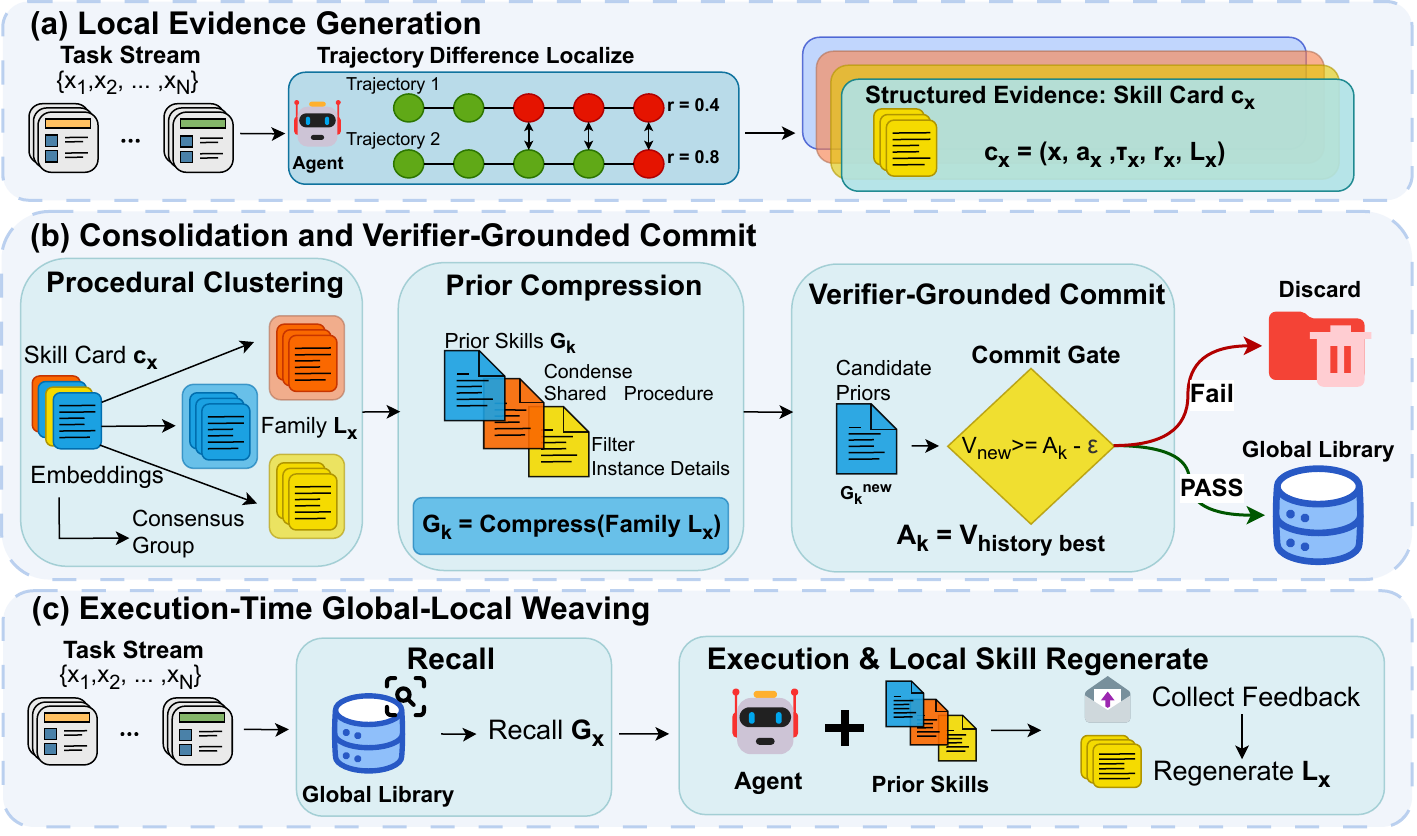}
\caption{Overview of GLoW. Each task's local skill is encapsulated as a skill card; cards are aggregated into procedural families and compressed into candidate global priors; candidates are committed only through a verifier-grounded gate on real downstream execution; at execution time, the frozen prior is woven with a freshly generated local skill.}
\label{fig:overview}
\end{figure*}

Building on this observation, we propose \textbf{SkillGLoW} (Global--Local Weave; GLoW), a layered skill-organization framework (Figure~\ref{fig:overview}). GLoW groups execution evidence into procedural families and splits skills into two layers. The global layer compresses each family's shared procedure into a stable prior that guides search; the local layer is regenerated for each task from its own execution feedback, supplies the instance detail the prior cannot carry, and is what the next round consolidates. The two are combined at solving time and separated at consolidation time. The global prior stays frozen during a solving episode and is revised only offline, and a revision is committed only when it does not degrade the deployed library in real execution.

We evaluate GLoW across three models on four benchmarks: mathematical reasoning, terminal automation, software repair, and embodied control, giving 12 continual-improvement runs in which a multi-setting protocol separates the contributions of the frozen prior and within-task regeneration. The committed priors improve over the no-skill baseline by 17.2 points on average, with positive gains in all 12 runs, while the deployed library holds one prior per procedural family rather than one per task, $3.6\times$ more compact than the per-task pool.
Our contributions are as follows:
\begin{itemize}
\item We measure all three prior organizations---the single document, the per-task pool, and the pool with retrieval---on the same streams (\S\ref{sec:ablation}); each gains less than family consolidation, identifying the missing unit of reuse: the solving procedure shared by a cluster of tasks.
\item We propose GLoW, which adopts the procedural family as this unit
and gates long-term updates by real execution (\S\ref{sec:formulation}--\S\ref{sec:weave}).
\item We show that, across the same runs, the frozen priors raise the
mean success rate on unseen embodied tasks from 73.9\% to 83.9\%, indicating that what is consolidated is reusable procedure rather than
task memory.
\end{itemize}

\section{Related Work}

\paragraph{Skill construction and optimization.}
Agent skills are procedural knowledge reused at inference time
\citep{toolformer,skillsbench,agentskillssurvey}. An earlier line has
agents discover them as executable code through environment
exploration \citep{voyager,skillweaver}; the systems closest to ours
write them as text. AutoSkill mines candidates from long-term
conversations under a judge that governs additions and deletions
\citep{autoskill}, while Trace2Skill \citep{trace2skill} and, earlier,
ExpeL \citep{expel} pool many trajectories before distilling, since
single-trajectory summaries overfit. SkillOpt trains one skill
document as the external state of a frozen agent, admitting an edit
only when it improves performance on a held-out split, and scopes it to a
single domain \citep{skillopt}. Skills of this kind belong to a broader
line that treats text as an optimizable parameter, whether through
prompt optimizers \citep{textgrad,dspy,opro}, self-feedback refinement
\citep{selfrefine}, or context compression \citep{acon}. In that line
the optimized object need only stay valid on one distribution; ours has
to hold across a family of tasks. These methods ask how to produce a better skill; we ask at what
granularity generated skills should be maintained.

\paragraph{Skill organization, transfer, and abstraction.}
Most systems store skills in a single layer, either one continually
rewritten document \citep{skillopt} or a flat retrievable library
\citep{autoskill,memento}, and lean on retrieval and governance: ReMe
deduplicates with a judge and prunes low-utility entries
\citep{reme}. The same granularity question runs through
procedural-memory work, whose stored unit ranges over reusable
workflows \citep{awm}, distilled procedures \citep{memp},
reasoning traces \citep{reasoningbank}, cross-domain experience entries
\citep{agentkb}, and test-time scratchpads \citep{dyncheatsheet}. That
evidence comes largely from structurally similar tasks; relax the
assumption and cross-domain retrieval induces negative transfer, with
abstract memories traveling better than task-specific detail
\citep{memorytransfer}. XSkill is closest architecturally, pairing an
action-level experience stream with a task-level skill stream, both
grounded in visual observations for multimodal tool use
\citep{xskill}. Its local component is retrieved from past records and
adapted; ours comes from the current task's own feedback and is never
written back.

\paragraph{Lifecycle and multi-skill maintenance.}
A third line maintains the skill system under long-term use. SkillOS
casts self-evolution as library governance, freezing the executor and
learning when the library should change \citep{skillos}. SkillGraph
adds prerequisite structure to retrieval \citep{skillgraph}; others
extend governance to runtime loops and multi-agent settings
\citep{museautoskill,skillmas,skillflow,bottomup}. All relocate the difficulty to when and how entries
are edited, and all need a utility scorer, learned or hand-set. GLoW
keeps no such policy: the global layer is re-derived each round from
that round's local evidence, so retention falls out of consolidation,
and acceptance turns only on measured downstream execution.

\section{Method}
\subsection{Problem Formulation}
\label{sec:formulation}
We define a \emph{skill} as a natural-language procedure inserted into
the agent's context. Given a frozen agent $\pi$, a task $x$, a skill
context $s$, and an execution harness $h$, one execution returns a
trajectory and a verifier score
$(\tau_x(s),\,r_x(s))=h(\pi,x,s)$ with $r_x(s)\in[0,1]$. That is,
changing $s$ is our only handle on the frozen agent's behavior, and
$r_x$ is the only trustworthy measure of effect.
GLoW maintains a committed global prior library
$\mathcal{G}=\{G_{\mathrm{base}},G_1,\ldots,G_K\}$, where each $G_k$ carries the
procedural structure shared by one family of tasks and $K$ is far
smaller than the number of historical tasks. For a task $x$, the
system recalls the relevant prior from $\mathcal{G}$:
\begin{equation}
G_x = \operatorname{Recall}(\mathcal{G},x) = G_{\mathrm{base}} \oplus G_{k(x)},
\label{eq:recall}
\end{equation}
where $\oplus$ denotes context concatenation, $G_{\mathrm{base}}$ is
injected unconditionally, and $G_{k(x)}$ is the family prior recalled
by similarity (\S\ref{sec:weave}). A Localize module generates a
task-local skill $L_x=\operatorname{Localize}(\Delta\tau_x,\,r_x)$,
where $\Delta\tau_x$ collects the differences between
successive trajectories of $x$, and $r_x$ denotes the corresponding verifier
scores; $L_x$
depends only on the current task's own execution feedback, queries no
historical library, and is not written into the long-term library.
GLoW's objective is to construct and maintain $\mathcal{G}$ over the
task stream so that the expected execution value of the combined
context is maximized:
\begin{equation}
\max_{\mathcal{G}}\; \mathbb{E}_{x\sim\mathcal{D}}\Bigl[\, r_x\bigl(\operatorname{Recall}(\mathcal{G},x)\oplus L_x\bigr) \Bigr],
\label{eq:objective}
\end{equation}
where $\mathcal{D}$ denotes the task distribution of the stream. Library
size is set by the family structure rather than by the stream, since
consolidation maintains one prior per family (\S\ref{sec:cluster}), so
$|\mathcal{G}|$ tracks the number of distinct procedures found, not the
number of tasks seen.
Optimizing Eq.~\eqref{eq:objective} is hard because $r_x$ is a black
box, the granularity of $\mathcal{G}$ is unknown, and language
compression is lossy; \S\ref{sec:local}--\S\ref{sec:weave} address these
in turn (Algorithm~\ref{alg:glow}).

\begin{algorithm}[!t]
\caption{GLoW: one pass over a task stream}
\label{alg:glow}
\begin{algorithmic}[1]
\REQUIRE agent $\pi$, harness $h$, stream $\mathcal{D}$, rounds $T$, sub-rounds $J$
\ENSURE committed prior library $\mathcal{G}$
\STATE $\mathcal{G} \leftarrow \{G_{\mathrm{base}}\}$
\FOR{$t = 1$ to $T$}
  \STATE \textbf{\textsc{Stage 1}} \emph{local evidence};\; $\mathcal{C} \leftarrow \emptyset$
  \FOR{$j = 1$ to $J$}
  \FORALL{task $x \in \mathcal{D}$}
    \STATE $G_x \leftarrow \operatorname{Recall}(\mathcal{G}, x)$ \COMMENT{frozen for the episode}
    \STATE $(\tau_x, r_x) \leftarrow h(\pi, x, G_x)$
    \REPEAT
      \STATE $L_x \leftarrow \operatorname{Localize}(\Delta\tau_x, r_x)$ \COMMENT{no library lookup}
      \STATE $(\tau_x, r_x) \leftarrow h(\pi, x, G_x \oplus L_x)$
    \UNTIL{iteration budget exhausted}
    \STATE $\mathcal{C} \leftarrow \mathcal{C} \cup \{(x, a_x, \tau_x, r_x, L_x)\}$ \COMMENT{card; $L_x$ never enters $\mathcal{G}$}
  \ENDFOR
  \ENDFOR
  \STATE \textbf{\textsc{Stage 2}} \emph{consolidation}
  \STATE $\{\mathcal{C}_k\}_{k \le K} \leftarrow \textsc{ConsensusCluster}(\phi(\mathcal{C}))$ \COMMENT{auto-$K$}
  \STATE $\hat{G}_k \leftarrow \operatorname{Compress}(\mathcal{C}_k)$ for all $k \le K$ \COMMENT{de-instantiate}
  \STATE \textbf{\textsc{Stage 3}} \emph{admission}
  \STATE $\mathcal{G}' \leftarrow \{G_{\mathrm{base}}\} \cup \{\hat{G}_k\}_{k \le K}$ \COMMENT{the round's candidate revision}
  \STATE $\mathcal{G} \leftarrow \mathcal{G}'$ \textbf{if} $\textsc{Gate}(\mathcal{G}', \mathcal{G})$ accepts \COMMENT{one decision per round; \S\ref{sec:commit}}
\ENDFOR
\RETURN $\mathcal{G}$
\end{algorithmic}
\end{algorithm}

\subsection{Local Evidence and Skill Cards}
\label{sec:local}
A local skill is generated from trajectory differences between
multiple real executions of the same task. Localize compares what
changed between adjacent runs and, together with verifier feedback,
judges which changes truly advanced the task, retaining the
operational hypotheses that execution feedback has already supported
on the current task. To support subsequent clustering and
consolidation, GLoW encapsulates each task's local experience as a
\emph{skill card} $c_x=(x,a_x,\tau_x,r_x,L_x)$: the task instruction,
an abstract signature naming what the solution did, the trajectory
text, the verifier score, and the local skill. $\Delta\tau_x$ feeds
Localize and is not carried on the card.

\subsection{Procedural Clustering and Prior Compression}
\label{sec:cluster}
The basic unit of offline consolidation is a family of procedurally
similar local skills $L_x$. GLoW groups by ``how it is solved'' rather
than ``what the task is about''. Each card supplies four textual
views---the signature $a_x$, the instruction $x$, the skill text $L_x$,
and the trajectory $\tau_x$---encoded separately \citep{sbert} and
fused with fixed weights into $\phi(c)$. The signature carries the largest weight and
$L_x$ the smallest, so grouping follows the procedure rather than either
wording. Views and weights are fixed across benchmarks; values are
in the appendix. A single hierarchical clustering is sensitive to the linkage rule and to
the number of clusters \citep{ward}. GLoW therefore runs many of them,
varying both, and records how often each pair of tasks lands in the same
group. That co-occurrence frequency becomes a consensus similarity
\citep{consensus}, and the final families come from clustering on it. The number of clusters is
selected automatically by the Kneedle knee point \citep{kneedle} of the
silhouette curve \citep{silhouette}.
Each family is compressed into one candidate prior:
\begin{equation}
G_k^{(t)}=\operatorname{Compress}\bigl(\mathcal{C}_k^{(t)}\bigr).
\label{eq:compress}
\end{equation}
Whether a candidate prior enters the library is decided by the verifier-grounded gate of \S\ref{sec:commit}.
The goal of prior compression is de-instantiated induction over the
group's local skills: extract the recurring solving skeleton and
filter out bindings that hold only in a single instance. A generated
candidate prior retains only three kinds of content: applicability
conditions, the core solving procedure, and common failure modes; the filtered instance details are recovered by the local skill $L_x$ that is freshly regenerated at execution time.

\subsection{Verifier-Grounded Commit}
\label{sec:commit}
A candidate prior is not written into the library directly:
compression can widen a rule or drop a constraint, and the text offers
no way to tell. Admission therefore rests on measured execution. The
execution value of deploying pure global priors on a task set $D$ is
\begin{equation}
V(\mathcal{G};D)=\frac{1}{|D|}\sum_{x\in D} r_x\bigl(\operatorname{Recall}(\mathcal{G},x)\bigr),
\label{eq:value}
\end{equation}
where the argument to $r_x$ is the recalled prior alone, with no $L_x$
appended, so that the measured value is attributable to the prior
rather than to task-local adaptation.
A round's candidates come from three routes. The first is fresh
compression of that round's local skills (\S\ref{sec:cluster}). The
second is append-only repair: where the deployed prior lost ground,
scope-restricted guards distilled from the degraded tasks' $L_x$ are
appended, leaving the verified structure intact. The third
carries forward, for each family, whichever version has scored best so
far. Every route yields one prior per family, and the priors of a round
together form a candidate revision $\mathcal{G}^{(v)}$, scored under
Eq.~\eqref{eq:value}. The gate keeps the best:
\begin{equation}
\begin{split}
V^{\mathrm{real}} &= \max_{v}\, V\bigl(\mathcal{G}^{(v)};D\bigr),\\
\text{commit} &\iff V^{\mathrm{real}} \;\ge\; A-\epsilon.
\end{split}
\label{eq:gate}
\end{equation}
The winner is committed only when Eq.~\eqref{eq:gate} holds; otherwise
the library is left untouched. Admission is decided once per round over
the revision as a whole, not separately per family. The anchor
$A=\max\bigl(V^{\star},\,V^{\mathrm{ns}}\bigr)$ is the larger of the
standing library's value and the No-Skill baseline, and
$\epsilon=0.02$ absorbs fluctuation. $A$ is floored further at the
highest value any measurement in the previous round reached, so a
revision cannot pass by clearing the standing commit alone. The maximum therefore selects among
candidate texts, not among repeated measurements of one, and since the
anchor is itself measured, a round can change the library only by
performing at least on par with the standing result, up to measurement noise ($\epsilon=0.02$).

\begin{table*}[!t]
\centering
\small
\setlength{\tabcolsep}{4pt}
\renewcommand{\arraystretch}{1.15}
\begin{tabular*}{\textwidth}{@{\extracolsep{\fill}}l cc cc cc c cc @{}}
\toprule
& \multicolumn{2}{c}{\textbf{TBP}$\uparrow$}
& \multicolumn{2}{c}{\textbf{SWE}$\uparrow$}
& \multicolumn{2}{c}{\textbf{ALFWorld}$\uparrow$}
& \textbf{LMB}$\uparrow$
 & \multicolumn{2}{c}{\textbf{Avg.}$\uparrow$} \\
\cmidrule(lr){2-3}\cmidrule(lr){4-5}\cmidrule(lr){6-7}\cmidrule(lr){8-8}\cmidrule(lr){9-10}
\textbf{Method} & hard & soft & hard & soft & hard & soft
 & hard\,$=$\,soft & hard & soft \\
\midrule
\multicolumn{10}{l}{\textit{\textbf{DeepSeek-V4-Pro}}} \\
\midrule
No-Skill & 34.4\nosub & 68.9\nosub & 45.0\nosub & 47.5\nosub & 57.1\nosub & 67.9\nosub & 13.2\nosub & 37.4\nosub & 49.4\nosub \\
SkillOpt$^{\dagger}$ & 43.8\gainsub{9.4} & 64.8\dropsub{4.1} & 60.0\gainsub{15.0} & 64.8\gainsub{17.3} & \textbf{90.5}\gainsub{33.4} & \textbf{94.4}\gainsub{26.5} & 22.6\gainsub{9.4} & 54.2\gainsub{16.8} & 61.7\gainsub{12.3} \\
GLoW (Global) & \textbf{50.0}\gainsub{15.6} & 74.9\gainsub{6.0} & \textbf{70.0}\gainsub{25.0} & 72.4\gainsub{24.9} & 83.3\gainsub{26.2} & 88.1\gainsub{20.2} & 37.7\gainsub{24.5} & \textbf{60.3}\gainsub{22.9} & 68.3\gainsub{18.9} \\
GLoW (Global+Local) & 43.8\gainsub{9.4} & \textbf{81.0}\gainsub{12.1} & \textbf{70.0}\gainsub{25.0} & \textbf{72.5}\gainsub{25.0} & 81.0\gainsub{23.9} & 81.0\gainsub{13.1} & \textbf{39.6}\gainsub{26.4} & 58.6\gainsub{21.2} & \textbf{68.5}\gainsub{19.1} \\
\midrule
\multicolumn{10}{l}{\textit{\textbf{MiniMax-M3}}} \\
\midrule
No-Skill & 40.6\nosub & 77.1\nosub & 35.0\nosub & 39.8\nosub & 76.2\nosub & 90.5\nosub & 22.6\nosub & 43.6\nosub & 57.5\nosub \\
SkillOpt$^{\dagger}$ & 43.8\gainsub{3.1} & 76.2\dropsub{0.9} & 55.0\gainsub{20.0} & 55.0\gainsub{15.2} & \textbf{95.2}\gainsub{19.0} & \textbf{97.6}\gainsub{7.1} & 28.3\gainsub{5.7} & 55.6\gainsub{12.0} & 64.3\gainsub{6.8} \\
GLoW (Global) & \textbf{53.1}\gainsub{12.5} & \textbf{86.7}\gainsub{9.6} & 60.0\gainsub{25.0} & 62.5\gainsub{22.7} & 92.9\gainsub{16.7} & 94.5\gainsub{4.0} & 28.3\gainsub{5.7} & 58.6\gainsub{15.0} & 68.0\gainsub{10.5} \\
GLoW (Global+Local) & \textbf{53.1}\gainsub{12.5} & 83.6\gainsub{6.5} & \textbf{65.0}\gainsub{30.0} & \textbf{67.5}\gainsub{27.7} & 90.5\gainsub{14.3} & 93.7\gainsub{3.2} & \textbf{35.8}\gainsub{13.2} & \textbf{61.1}\gainsub{17.5} & \textbf{70.2}\gainsub{12.7} \\
\midrule
\multicolumn{10}{l}{\textit{\textbf{GPT-5.4-mini}}} \\
\midrule
No-Skill & 28.1\nosub & 75.6\nosub & 35.0\nosub & 37.4\nosub & 42.9\nosub & 60.9\nosub & 15.1\nosub & 30.3\nosub & 47.2\nosub \\
SkillOpt$^{\dagger}$ & 40.6\gainsub{12.5} & 80.0\gainsub{4.4} & 40.0\gainsub{5.0} & 42.4\gainsub{5.0} & 61.9\gainsub{19.0} & 74.2\gainsub{13.3} & 13.2\dropsub{1.9} & 38.9\gainsub{8.7} & 52.5\gainsub{5.2} \\
GLoW (Global) & \textbf{50.0}\gainsub{21.9} & 83.0\gainsub{7.4} & 45.0\gainsub{10.0} & 47.4\gainsub{10.0} & 61.9\gainsub{19.0} & \textbf{77.8}\gainsub{16.9} & \textbf{18.9}\gainsub{3.8} & 43.9\gainsub{13.6} & 56.8\gainsub{9.6} \\
GLoW (Global+Local) & 46.9\gainsub{18.8} & \textbf{85.8}\gainsub{10.2} & \textbf{50.0}\gainsub{15.0} & \textbf{55.7}\gainsub{18.3} & \textbf{66.7}\gainsub{23.8} & 76.0\gainsub{15.1} & \textbf{18.9}\gainsub{3.8} & \textbf{45.6}\gainsub{15.3} & \textbf{59.1}\gainsub{11.9} \\
\bottomrule
\end{tabular*}

\caption{Main results (\%, 12 continual-improvement runs). Subscripts give
the change over that model's No-Skill row, computed before rounding
(\gainval{}\,gain,
\dropval{}\,loss; the arrow, not the color, carries the sign). LMB is
binary multiple-choice, so hard $=$ soft. Best per column within a
model block in
\textbf{bold}. $^{\dagger}$Run protocol is given in \S\ref{sec:main}.}
\label{tab:main}
\end{table*}

\subsection{Execution-Time Global--Local Weaving}
\label{sec:weave}
At test time, the committed global prior library stays frozen. For a
new task $x$, Recall implements Eq.~\eqref{eq:recall} in two levels:
the base prior is always present; on the family-prior side, the system
builds an embedding for the full text of each committed prior
\citep{qwen3embed}, embeds
the task instruction as the query, takes the cosine-similarity top-1,
and injects it only when similarity exceeds a fixed threshold shared
across benchmarks, otherwise falling back to $G_{\mathrm{base}}$ alone
(fail-closed). Recall completes before the task starts and stays fixed
throughout the task's execution.
The agent then executes under $G_x$ and collects feedback from the
current task. Localize generates a fresh local skill $L_x$ as in
\S\ref{sec:local}. In subsequent solving, the agent uses
$G_x \oplus L_x$ as the skill context and keeps updating $L_x$. This
$L_x$ is staged as local evidence of the current sub-round.
After the round, these $L_x$ feed consolidation and admission
(Alg.~\ref{alg:glow}, Stages 2--3).

\section{Experiments}

\subsection{Setup}

\paragraph{Benchmarks.}
We use four benchmarks, ordered from low to high cross-task
procedure sharing: Terminal-Bench-Pro (TBP) \citep{tbpro}, SWE-bench Verified
\citep{jimenez2024swebench} (solved through an agentic harness
\citep{sweagent}), ALFWorld \citep{shridhar2021alfworld}, and
LiveMathematicianBench (LMB) \citep{lmb}. In the first two each task is
an independent repository or terminal scenario with a distinct
solution, which is the long-horizon, heterogeneous-solution regime this
paper targets. The four streams contain 32, 20, 42, and 53 tasks, respectively.

\paragraph{Models and settings.}
Each run uses one of three models as its single frozen base model for
solving, extraction, and compression: DeepSeek-V4-Pro, MiniMax-M3, and
GPT-5.4-mini, giving 12 runs. Every setting is a measurement point
inside one continual run; they differ only in what occupies the skill
context. \textbf{No-Skill}
leaves it empty and establishes the baseline. \textbf{Local-only} carries
the skill the task regenerated from its own feedback and recalls
nothing, isolating within-task adaptation. \textbf{Global-only}
carries the recalled prior with regeneration disabled, isolating what
consolidation has taken from the local skills.
\textbf{Global+Local} carries both. A fifth setting,
\textbf{Base-only}, replaces the library with a single compressed
document and appears only in the ablations. All five are
configurations of our own system.
Hard and soft are readings of the
verifier score $r_x$ (\S\ref{sec:formulation}): \emph{hard} is each benchmark's
own all-or-nothing flag. \emph{Soft} is the partial-credit signal the
benchmark defines, or one of ours where it defines none; on software
repair it is multiplicative, so breaking a passing test zeroes the
score. Hard is sparse on long-horizon
tasks; soft records progress short of it. Per-benchmark definitions are in the appendix. Recall uses Qwen3-Embedding-8B under 4-bit quantization
\citep{qwen3embed}, with a cosine-similarity threshold of $0.45$ across benchmarks;
below it a task receives no prior. Every main-table cell is a single run
at provider defaults (temperature $0.7$); seeds fix the splits and the
task order. The
held-out software-repair numbers average three trials. The base models are hosted APIs; recall runs on one GPU (16\,GB VRAM) with 32\,GB host RAM on Linux (WSL2).
\subsection{Main Results}
\label{sec:main}

Both GLoW settings improve on No-Skill in every one of the 12 runs
(Table~\ref{tab:main}; Wilcoxon signed-rank $p=0.000488$, the smallest
value attainable at $n=12$): the frozen priors gain $17.2$/$13.0$ points (hard/soft) on
average, and $18.0$/$14.6$ once local regeneration is added.
The gain holds up at the software-repair end, where each task is a
separate repository and the two existing organizations have the least
to work with: the largest gain anywhere in
Table~\ref{tab:main}, on both readings, falls there (MiniMax-M3,
$+30.0$/$+27.7$ with regeneration), and DeepSeek-V4-Pro adds
$+25.0$/$+25.0$, whereas the mathematical end, whose tasks share the
most procedural structure, is where two of three models gain least
($+3.8$ and $+5.7$). That is the ordering the framework predicts,
since family granularity pays off precisely where a single document
has no dominant procedure to summarize.

SkillOpt \citep{skillopt} tests that ordering. We run the authors' own
implementation with their internal hyper-parameters untouched, match only
the launch protocol (task stream, trials, decoding) to ours, and score it
through the same adapters as every other row, so all numbers share one
source; its entry is the peak over training epochs, matching
GLoW's peak-over-rounds reading (appendix). Across all 21 cells, GLoW leads on 15, SkillOpt on 4 and two tie.
SkillOpt's four are all ALFWorld, the one benchmark whose tasks share a
single action space: where one skeleton covers the whole set, a single
document is already the right granularity. Elsewhere the margin runs the
other way ($+8.3$ on TBP, $+6.7$ on SWE, and $+6.9$ on LMB, hard, averaged over
the three models).

On unseen tasks the prior alone carries the gain (appendix). The accumulation is also achieved on a bounded library; across
the 12 runs the committed library is smaller than the per-task skill
pool by a factor of $3.6$.

\subsection{Component Ablations}
\label{sec:ablation}

The two organizational extremes are a single merged document \citep{trace2skill} and a per-task entry library \citep{museautoskill,autoskill}. Table~\ref{tab:main} carries a published instance of the first; here we place our own library at each extreme instead, so that both ends are measured under identical conditions (Table~\ref{tab:ablation}): Base-only compresses the whole library into a single injected skill, and Local-only uses only the skill regenerated from each task's own feedback.

\begin{table}[!t]
\centering
\footnotesize
\setlength{\tabcolsep}{3pt}
\renewcommand{\arraystretch}{1.15}
\begin{tabular}{llccc}
\toprule
 &  & Base-only & Local-only & AWM \\
Bench & Model & {\scriptsize (w/o families)} & {\scriptsize (w/o global)} & {\scriptsize (flat\,+\,recall)} \\
\midrule
\multirow{3}{*}{TBP}
 & DeepSeek-V4-Pro & 25.1\dropsub{9.3} & 48.5\gainsub{14.1} & 37.5\gainsub{3.1} \\
 & MiniMax-M3 & 46.9\gainsub{6.3} & 50.0\gainsub{9.4} & 34.4\dropsub{6.2} \\
 & GPT-5.4-mini & 34.4\gainsub{6.3} & 46.9\gainsub{18.8} & 40.6\gainsub{12.5} \\
\midrule
\multirow{3}{*}{SWE}
 & DeepSeek-V4-Pro & 45.0\nosub & 57.5\gainsub{12.5} & 65.0\gainsub{20.0} \\
 & MiniMax-M3 & 30.0\dropsub{5.0} & 50.0\gainsub{15.0} & 30.0\dropsub{5.0} \\
 & GPT-5.4-mini & 30.0\dropsub{5.0} & 45.0\gainsub{10.0} & 40.0\gainsub{5.0} \\
\midrule
\multirow{3}{*}{ALFWorld}
 & DeepSeek-V4-Pro & 73.8\gainsub{16.7} & 73.8\gainsub{16.7} & 76.2\gainsub{19.1} \\
 & MiniMax-M3 & 78.6\gainsub{2.4} & 79.8\gainsub{3.6} & 85.7\gainsub{9.5} \\
 & GPT-5.4-mini & 50.0\gainsub{7.1} & 53.6\gainsub{10.7} & 45.2\gainsub{2.3} \\
\midrule
\multirow{3}{*}{LMB}
 & DeepSeek-V4-Pro & 20.8\gainsub{7.6} & 24.5\gainsub{11.3} & 13.2\nosub \\
 & MiniMax-M3 & 18.9\dropsub{3.7} & 32.1\gainsub{9.5} & 24.5\gainsub{1.9} \\
 & GPT-5.4-mini & 15.1\nosub & 14.2\dropsub{0.9} & 13.2\dropsub{1.9} \\
\midrule
Mean & --- & 39.1\gainsub{2.0} & 48.0\gainsub{10.9} & 42.1\gainsub{5.0} \\
\bottomrule
\end{tabular}
\caption{\textbf{Three alternative skill organizations, on the same 12
runs.} Hard score (\%); a subscript gives the change over that row's
No-Skill in Table~\ref{tab:main}, rendered as there, and a cell without one did
not move. AWM is a flat pool with retrieval.}
\label{tab:ablation}
\end{table}

\paragraph{Granularity (w/o families / w/o global).}
Removing the procedural-family layer collapses the system to one
extreme or the other, and the two fail in opposite ways. Base-only is
the weak form of the single-document organization, one compression
pass with no iteration. It averages $+2.0$ points, flips sign across
benchmarks ($-9.3$ to $+16.7$), and loses ground on 4 of the 12 cells.
SkillOpt is the state of the art for a single persistent document, and
clears Base-only on 10 of the 12 cells (Tables~\ref{tab:main},~\ref{tab:ablation}); the
distance between them measures optimization effort rather than
organization. Base-only succeeds consistently only on ALFWorld, for the
reason \S\ref{sec:main} gives.
Elsewhere a merged document yields generic
discipline the model has already internalized. On software repair it never helps: the
resolved count is flat for one model and lower for the other two, and an
independent benchmark of
authored skills reports 39 of 49 leaving the pass rate unchanged
\citep{swebskills}. Per-task skills carry the opposite defect. They
help, by $+10.9$ points, but the entry is built around the instance
(\S\ref{sec:analysis}), and the pool grows linearly with the stream;
appending such entries without curation is known to accumulate noise and
staleness rather than reuse \citep{reme,skillos}, and cross-task recall
over them transfers little \citep{memorytransfer}. The setting still
trails Global-only.
Adding retrieval to that pool does not close the gap. Workflows induced
in the style of AWM \citep{awm} from the same first-round evidence,
pooled and retrieved per task, average $+5.0$ points across the 12
cells, eight up, three down and one flat, while our own one round of family consolidation gives $+11.2$ (appendix).

\begin{figure}[!t]
\centering
\includegraphics[width=\columnwidth]{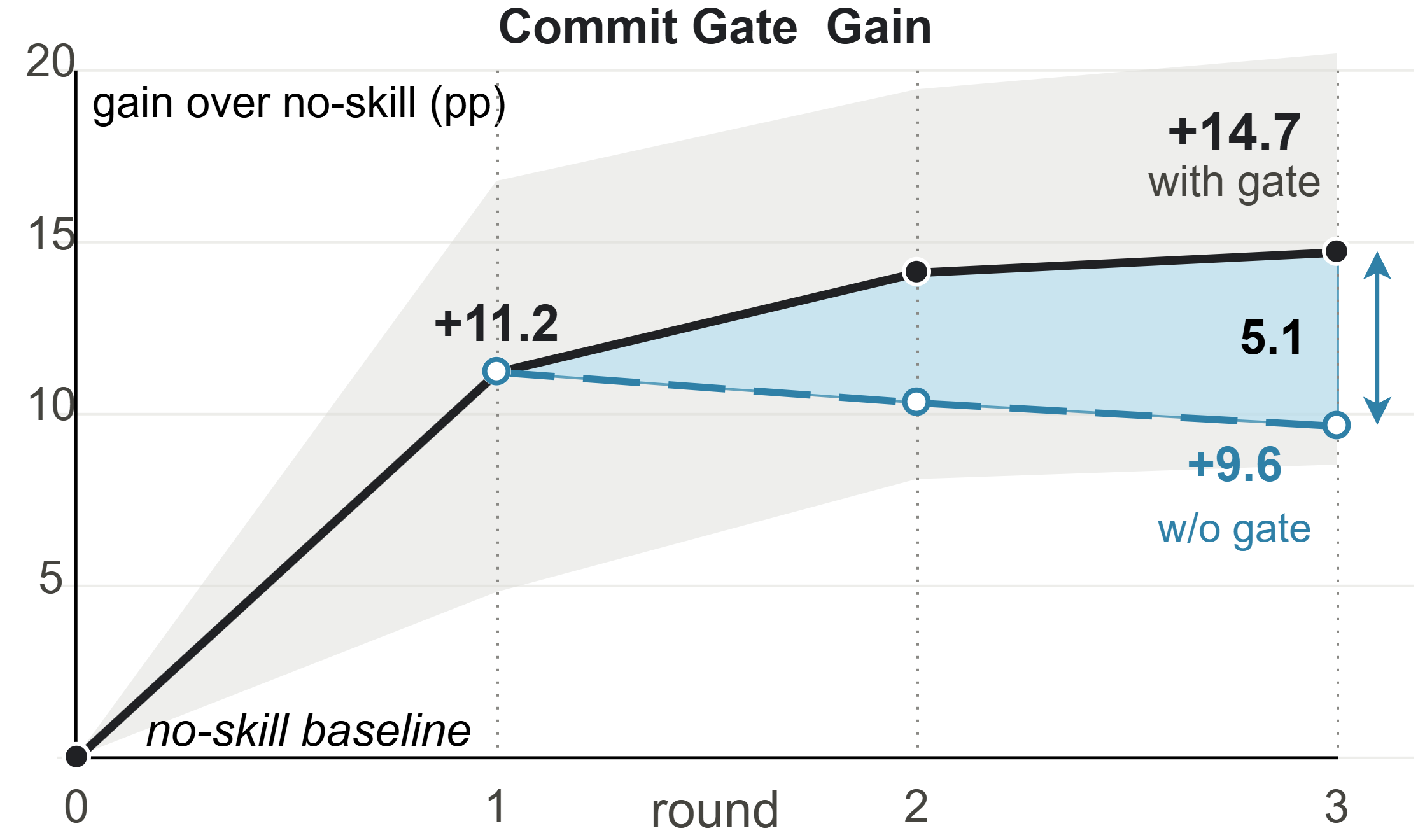}
\caption{Mean gain over No-Skill (hard, points), 12 runs; band is the
15th--85th percentile. \textbf{Solid}: what the gate admitted, held flat on
a rejection. \textbf{Dashed}: that round's candidate, admitted
unconditionally. Rounds are compared independently.}
\label{fig:gate}
\end{figure}

\paragraph{Admission (w/o gate).}
Ablating the gate (\S\ref{sec:commit}) requires no separate control run, as the
gate already evaluates every candidate under real deployment, so the
cost of an erroneous admission is measured rather than simulated
(Figure~\ref{fig:gate}). Across the 12 runs, the gate made 26 admission
decisions, accepting 19 and rejecting 7. Decisions depend on the
workload rather than the model. On terminal tasks, candidates for all three models
were admitted in both rounds; on software repair, all three were
admitted in the first round and rejected in the second. A cheaper
criterion would not reproduce these rejections, as 4 of the 7 would
have been admitted under the consolidation-time score alone. A representative
case is MiniMax-M3 on mathematical reasoning, whose round-2 candidate
scored $0.283$ on the soft signal the gate uses, exceeding the anchor,
yet achieved only $0.151$ in
deployment ($-13.2$ points, more than twice the largest run-to-run range
we measure). Averaged over the 12 runs,
the library the gate kept reaches $+14.7$ points by the final round,
while that round's candidates average $+9.6$ (Figure~\ref{fig:gate}).

\subsection{Analysis}
\label{sec:analysis}

\begin{table}[!t]
  \centering
  \footnotesize
  \setlength{\tabcolsep}{1pt}
  \renewcommand{\arraystretch}{1.15}
  \begin{tabular}{lcccc}
  \toprule
  Task class & Tasks & Priors & Solved & Fix\,/\,Break \\
   & {\scriptsize (dataset)} & {\scriptsize (skill num)}
   & {\scriptsize (no skill\,$\to$\, with global)} & \\
  \midrule
  Examine-in-light  & 7 & 1 & 4 $\to$ 7 & \gainval{3} \\
  Pick-and-place    & 7 & 3 & 7 $\to$ 7 & --- \\
  Clean-then-place  & 7 & 3 & 4 $\to$ 5 & \gainval{2}\,\dropval{1} \\
  Cool-then-place   & 7 & 2 & 5 $\to$ 6 & \gainval{1} \\
  Heat-then-place   & 7 & 3 & 5 $\to$ 7 & \gainval{2} \\
  Two-instance      & 7 & 3 & 7 $\to$ 7 & --- \\
  \midrule
  All six classes   & 42 & 15 & 32 $\to$ 39 & \gainval{8}\,\dropval{1} \\
  \bottomrule
  \end{tabular}
  \caption{Priors consolidated onto ALFWorld's six canonical task classes
  (MiniMax-M3). The classes are the dataset's own labels, produced
  independently of our algorithm. \emph{Solved}: No-Skill $\to$ round~2,
  in-sample. \emph{Fix\,/\,Break}: tasks that flipped $0\!\to\!1$ and
  $1\!\to\!0$; the arrow, not the color, carries the sign.}
  \label{tab:class-counts}
  \end{table}

\paragraph{Priors are organized by procedure.}
ALFWorld's own task labels give a partition we did not produce.
Consolidation returns 15 priors over 42 tasks: 13 sit wholly inside one
labeled class, and two straddle a pair and are counted by majority type
(Table~\ref{tab:class-counts}). No class is left uncovered and none is
invented. This is also why Base-only is reliably positive on this
benchmark alone (Table~\ref{tab:ablation}): here a shared skeleton
exists at the scale of the whole task set. Terminal-Bench-Pro's 32 tasks
each carry their own subject matter, yet compression still returns ten
procedure classes. Five read as \emph{binary} work, but the
byte-scanning prior helps only the two actually solved by scanning raw
bytes, and leaves the other three untouched. The appendix gives the full
terminal mapping.

\paragraph{What a prior keeps.}
Compression is meant to strip the instance and retain the procedure
(\S\ref{sec:cluster}), and the committed text is where that can be
checked. The local skill for the cell-phone instance of \texttt{examine-in-light}
spends 989 words on
seven steps built around a memorized location table: receptacles ranked
``soft furnishings $>$ dresser-top $>$ desk $>$ drawers \dots desk for
desklamp, bedside/nightstand for bedlamp.'' The
prior its seven-task class compresses into spends 440 words on four
steps, and the table is gone. Step~1 replaces it with a read: scan the
initial observation for the illuminator-class object, ``do not default
to \texttt{desk} \dots verify against the visible object list before any
\texttt{go to}.'' The rest is the chain that follows: go to the source
receptacle and take the target without pre-examining it; carry it to the
illuminator's receptacle without exploratory \texttt{examine} calls en
route; terminate with \texttt{use <illuminator>} and append nothing
after. Every pitfall in the chain guards one thing: turns spent on
probes that return no new information. That is what the seven tasks
share; which furniture holds what is regenerated per task.

\paragraph{An aligned class repairs failures without memorizing them.}
On Terminal-Bench-Pro (GPT-5.4-mini) the same accounting gives 9 repairs
against one pass-to-fail; ALFWorld's counts are in Table~\ref{tab:class-counts},
where the single regression falls inside one class and the gain
concentrates where a class had failures to repair. The three tasks fixed
in \texttt{examine-in-light} are an alarm clock, a book, and a cell
phone: one prior, three different objects. The reverse case is equally
visible. The largest family put eight Terminal-Bench-Pro
tasks together, among them a debugger session, a write-ahead-log
recovery, and a logic-gate CRC32 build, and took its name,
\texttt{segfault\_debugging\_with}, from one member. Its prior repaired
one of the eight and broke another, the only large family in the study with no
net gain. On the same 32 tasks a second model drew ten families instead,
none larger than five. Procedure sharing is therefore a property of how
the family is drawn rather than of the benchmark. Families are
re-derived from each round's own evidence, so a bad grouping is not
carried into the next.

\subsection{Transfer to Unseen Tasks}
\label{sec:transfer}

Does the library transfer, or has it memorized the stream? We inject the library the training stream produced, unmodified, into 60 unseen tasks of ALFWorld valid\_unseen, a split that shares task categories with the training tasks but no instances.

\begin{table}[!t]
\centering
\small
\setlength{\tabcolsep}{5pt}
\renewcommand{\arraystretch}{1.15}
\begin{tabular}{lcc}
\toprule
Model & No-Skill & Global \\
\midrule
DeepSeek-V4-Pro & 71.7 & \textbf{88.3}\gainsub{16.7} \\
GPT-5.4-mini    & 66.7 & \textbf{75.0}\gainsub{8.3} \\
MiniMax-M3      & 83.3 & \textbf{88.3}\gainsub{5.0} \\
\midrule
Mean            & 73.9 & \textbf{83.9}\gainsub{10.0} \\
\bottomrule
\end{tabular}
\caption{Transfer to unseen ALFWorld tasks (hard, \%; the last row
averages the three models; subscripts as in Table~\ref{tab:main}). All
three models improve.}
\label{tab:transfer}
\end{table}
All three models improve (Table~\ref{tab:transfer}), and the gain
is smallest for the model with the highest baseline, consistent with a
ceiling effect.

On software repair the same library lifts the resolve rate on 30 unseen instances from
40.0\% to 45.6\% (MiniMax-M3, mean of three trials), the workload
where transfer should be hardest, since every instance is a different
repository. For reference, \citet{memorytransfer} report a $3.7\%$
average gain for cross-domain memory transfer, in a setting that pools
heterogeneous domains rather than holding out instances of one
benchmark.

\section{Conclusion}

Neither a single global document nor a flat per-task library survives a
stream whose tasks each need a different solution. Experiments across 12
continual-improvement runs show the missing unit is the procedural
family: compress the local skills of each cluster into one
de-instantiated prior, regenerate instance detail per task rather than
storing it, and admit priors only through measured execution. This recipe gains 17.2
points over No-Skill on a library $3.6\times$ smaller than the per-task
pool, and transfers unmodified to unseen tasks (73.9\% to 83.9\%). Transfer was shown where task categories recur; whether a prior survives
a genuine domain change remains open. Because a prior is plain text, one model's library could in principle be handed to another; such cross-model inheritance remains untested. Both point toward a library any open-ended
workflow could accumulate, rather than one tied to a benchmark.

\bibliography{references}

\newpage
\appendix
\input{supplementary}

\end{document}

%% file: supplementary.tex
\title{Supplementary Document\\SkillGLoW: Procedural-Family Skill Consolidation
for Self-Improving Agents on Long-Horizon Task Streams}
\author{Anonymous Submission}
\affiliations{Paper under double-blind review}

\maketitle

\noindent
This document supplies the material the paper defers to. Section letters
(\S\S A--O), tables (Tables~A1--A28) and figures (Figures~A1--A3) are local to
this document; a table, figure, equation or \S-number without a letter points
into the main paper. Reviewers are not obliged to consult this document, and the
paper is self-contained without it. The method is \emph{SkillGLoW} in the
title and \emph{GLoW} for short throughout.

\appendix
\renewcommand{\thetable}{A\arabic{table}}
\renewcommand{\thefigure}{A\arabic{figure}}
\setcounter{table}{0}
\setcounter{figure}{0}

\section{Metric Definitions}
\label{app:metrics}

The \emph{hard} metric is the all-or-nothing flag the benchmark ships with.
The \emph{soft} metric is the benchmark's own continuous partial credit;
ALFWorld has no official partial credit, so its soft score is derived by our
verifier.

\begin{apptab}
\begin{tabular}{@{} p{0.27\columnwidth} p{0.20\columnwidth} p{0.43\columnwidth}@{}}
\toprule
\textbf{Benchmark} & \textbf{hard} & \textbf{soft} \\
\midrule
Terminal-Bench-Pro
  & all \idt{pytest} tests pass
  & the fraction of \idt{pytest} tests passed; ships with the benchmark \\
SWE-bench Verified
  & binary \idt{resolved}
  & a multiplicative gate on the fail-to-pass and pass-to-pass rates \\
ALFWorld
  & binary success flag
  & the larger of the subgoal completion ratio and the TextWorld intermediate
    reward; derived here \\
Live\-Mathematician\-Bench
  & binary multiple choice
  & $\equiv$ hard \\
\bottomrule
\end{tabular}
\end{apptab}
\apptabcap{hard and soft, one benchmark at a time.}{tab:a-metrics}

Table~\ref{tab:a-metrics} gives them one benchmark at a time. SWE's soft is a
multiplicative gate: if either side is zero the soft score is
zero, so breaking a test that used to pass is never recorded as partial
success.

The four soft metrics live on different scales, so the \emph{Avg.} soft column
of \btab{tab:main} averages across read-outs and should be read only as
relative movement across settings within one row; the per-benchmark columns are
the comparable ones. The hard columns are unaffected.

The commit gate runs on soft while the headline reports hard. In
$V(\mathcal{G};D)=\frac{1}{|D|}\sum_x r_x$ of \bsec{sec:commit}, $r_x$ is each
benchmark's soft. On long-horizon tasks hard is too sparse --- the four
benchmarks' No-Skill hard lies between 13.2\% and 76.2\% --- and using it as
the gate signal would frequently give the wrong verdict, because a single task
flipping can decide it. Accordingly the two scores in \bsec{sec:ablation},
candidate $0.283$ against deployed $0.151$, are both soft.

\section{Multi-View Fusion and Consensus Clustering}
\label{app:fusion}

Each local skill card is encoded as up to five view vectors; each is
$L_2$-normalised and the weighted sum gives $\phi(c)$. The body describes four
views, with the signature taking two of them --- one compact and one verbose
granularity, encoded separately.

\begin{apptab}
\begin{tabular}{@{}c l p{0.52\columnwidth}@{}}
\toprule
& \textbf{Channel} & \textbf{Content} \\
\midrule
S & Signature, compact & the 2--5-word abstract operation name \\
L & Signature, verbose & a 10--15 word description of the procedure \\
I & Task instruction   & the task text as given \\
T & Trajectory text    & the goal and actions of the execution plan and the
                         checks and observations of the plan trace,
                         $\approx$500 characters \\
F & Full local skill   & the skill description concatenated with every
                         section, truncated at 8000 characters \\
\bottomrule
\end{tabular}
\end{apptab}
\apptabcap{The five view channels.}{tab:b-channels}

The five channels are listed in Table~\ref{tab:b-channels}. The tier is decided
at run time by which fields the round's cards actually carry, and the twelve
runs trigger only the two tiers of Table~\ref{tab:b-weights}.

\begin{apptab}
\begin{tabular}{@{}l p{0.30\columnwidth} rrrrr@{}}
\toprule
\textbf{Tier} & \textbf{Trigger}
 & \textbf{S} & \textbf{L} & \textbf{I} & \textbf{T} & \textbf{F} \\
\midrule
Five-view & any card carries a trajectory
 & 0.30 & 0.20 & 0.20 & 0.20 & 0.10 \\
Four-view & no trajectory, full skill present
 & 0.25 & 0.25 & 0.25 & \naval & 0.25 \\
\bottomrule
\end{tabular}
\end{apptab}
\apptabcap{The two weight tiers actually deployed. A dash means the channel is
absent in that tier.}{tab:b-weights}

The sweep behind the four-view values is Table~\ref{tab:b-sweep}. Both tiers
share the property that the signature channels together carry the
most weight, 0.50 in each, and the full skill text the least --- consistent
with the body's ``grouping follows the procedure rather than either wording''.
A missing channel falls back to that card's task instruction; the weights
therefore still sum to one. Across rounds the vectors accumulate under an exponential
moving average with $\alpha=0.2$. Both tiers' weights are hard-coded constants,
unadjusted across four benchmarks and three models.

The five-view tier is the global optimum of one five-view grid sweep at step
$0.1$. The four-view equal weighting comes from a separate sweep at the same
granularity, scored by ARI against the dataset's labels; the parenthesised
value is the number of clusters selected automatically.

\begin{apptabsm}
\begin{tabular}{@{}l r@{\ }l r@{\ }l r@{}}
\toprule
\textbf{Weights S/L/I/F}
 & \multicolumn{2}{c}{\textbf{ALFWorld ARI}}
 & \multicolumn{2}{c}{\textbf{Second domain}}
 & \textbf{Mean} \\
\midrule
0.40 / 0.40 / 0.20 / \naval  & 0.799 & (6)  & 0.557 & (15) & 0.678 \\
0.40 / 0.20 / 0.20 / 0.20    & 0.799 & (6)  & 0.520 & (6)  & 0.660 \\
0.30 / 0.20 / 0.20 / 0.30    & 0.799 & (6)  & 0.690 & (13) & 0.745 \\
0.30 / 0.30 / 0.10 / 0.30    & 0.799 & (6)  & 0.671 & (11) & 0.735 \\
\textbf{0.25 / 0.25 / 0.25 / 0.25}
                             & \textbf{0.878} & \textbf{(7)}
                             & \textbf{0.733} & \textbf{(16)}
                             & \textbf{0.806} \\
0.50 / 0.20 / 0.10 / 0.20    & 0.799 & (6)  & 0.448 & (17) & 0.624 \\
\naval\ / 0.40 / 0.20 / 0.40 & 0.775 & (6)  & 0.592 & (21) & 0.684 \\
\bottomrule
\end{tabular}
\end{apptabsm}
\apptabcap{The four-view weight sweep. The second domain is a
spreadsheet-manipulation benchmark and is not one of our four; of those, only
ALFWorld took part in this selection, and the other three inherited the result
directly.}{tab:b-sweep}

The cosine similarity matrix $S$ is min-max normalised to $[0,1]$ and the
distance is $D=1-S$. There are nine base partitions: average and complete
linkage on $D$, and Ward linkage on $S$, each run for
$K\in\{K_0-1,\,K_0,\,K_0+1\}$, where $K_0$ is a seed value from a first pass. The consensus matrix is the co-assignment
frequency
\begin{equation}
\mathrm{CO}_{ij}=\frac{1}{9}\sum_{p=1}^{9}
  \mathbf{1}\bigl[\ell_p(i)=\ell_p(j)\bigr],
\label{eq:consensus}
\end{equation}
and the final families come from one more average-linkage clustering on
$1-\mathrm{CO}$.

$K$ is the Kneedle knee of the silhouette curve, searched over
$K\in[3,\ \min(n-1,\ \lfloor n/2\rfloor+2)]$: both axes are normalised and the
$K$ with the largest perpendicular distance from the chord joining the
endpoints is taken. The criterion has no tunable parameter and prefers fewer
clusters than a plain \emph{argmax}. The $K$ values in
\app{app:family-structure} are produced this way.

\section{Run Protocol of the Compared Methods}
\label{app:baselines}

\subsection{SkillOpt}
\label{app:skillopt}

The SkillOpt row of \btab{tab:main} comes from the official implementation,
\idt{microsoft/\ab SkillOpt}, not from a reimplementation of our own. Apart from the
alignments below, its internal hyperparameters are the authors' defaults:
learning rate 4, minibatch 8, four epochs, batch 40, with the skill length
budget and all prompts untouched.

\begin{apptab}
\begin{tabular}{@{}p{0.19\columnwidth} p{0.40\columnwidth} p{0.24\columnwidth}@{}}
\toprule
\textbf{Aligned item} & \textbf{Our reference run} & \textbf{SkillOpt baseline} \\
\midrule
Task set & train 53\,/\,42\,/\,32\,/\,20 & the same batch \\
Model    & one model does solving, extraction and compression & same \\
Trials per task & 1 & 1 \\
thinking & off for LMB, on for the other three & same \\
Max interaction turns & 5\,/\,30\,/\,15\,/\,25 & same \\
Scoring  & our evaluation adapter & same \\
\bottomrule
\end{tabular}
\end{apptab}
\apptabcap{Item-by-item alignment between our reference runs and the SkillOpt
baseline. Every item that could move the comparison --- task set, base model,
trials per task, thinking, turn limit and scoring --- is held identical. Two
further items --- the initial skill and the optimiser model --- are not listed
in the table and are settled in the text below. Every slash-separated
quadruple is in the order LMB\,/\,ALFWorld\,/\,TBP\,/\,SWE.}{tab:c-skillopt-align}

Table~\ref{tab:c-skillopt-align} lists the alignment item by item, and the two
it leaves to the text are settled the same way on both sides --- each departs
from SkillOpt's official default, and each does so in the direction that
removes an asymmetry rather than one that creates one. The official
release provides hand-written initial skills for LMB and ALFWorld
only. Keeping them would leave the four benchmarks at different starting points
and would mix the contribution of the human prior into the same number as the
optimisation loop, so all four use one neutral 99-byte skeleton and the gain is
attributable entirely to the optimisation loop. That choice was verified: on
MiniMax-M3\,$\times$\,ALFWorld the blank start and the official hand-written
skill give digit-for-digit identical weighted scores across all three epochs,
so the hand-written document's net contribution in that cell is 0.0\,pp and the
$+19.0$\,pp is produced by the optimisation loop.

The original method allows the optimiser to be a stronger model distilling into
the target model. Our own pipeline has no such channel --- one frozen model does
solving, extraction and compression --- so we switch it off on SkillOpt too,
putting optimiser and solver on the same base model. What remains is a
comparison between two ways of organising skills for one and the same model,
with no distillation on either side.

All four benchmarks go through our evaluation adapter, so hard and
soft carry the same meaning as in every other table. SkillOpt's native
environments cover only ALFWorld and LMB, and those environments' read-outs are
not comparable with ours: native ALFWorld hard-codes soft as 1 on success, and native LMB
brings its own prompt and choice shuffling and returns $0.077$ on the same 53
tasks where our No-Skill baseline is $0.226$. Once both sides go through our
adapter, SkillOpt's score on that benchmark is higher than its own native
environment can produce.

SkillOpt splits each epoch into a 40-task batch and a remainder batch, so a
full pass exists only after merging an epoch's batches with weights; the figure
in the body is that epoch-weighted merged train peak.

\subsection{AWM}
\label{app:awm}

AWM induces, from experience, workflows that are shared across tasks, and is
conceptually the closest route to a procedural family prior. It is the AWM column of
\btab{tab:ablation}.

We use the offline setting (Wang et al., arXiv:2409.07429): given a batch of
already-attempted tasks, one LLM call extracts the sub-procedures that recur
across them as workflows, and the result is prepended whole to the agent's
prompt. The official repository ships only WebArena and Mind2Web branches, but
the method carries no web-specific assumption. Two decisions were needed for
the port. The induction instruction is used verbatim from the official
repository, read from it at run time. The one-shot example, by contrast, is
replaced by one of the same structure drawn from the target benchmark's own
action vocabulary, because the
official 138 lines of WebArena click actions would push the model to write
click-shaped workflows for an environment that has no clicking.

The trajectories that induction uses are rollouts under the No-Skill condition:
the first step of a SkillOpt run started from a blank skill, taken before the
optimiser has made any update and with an empty skill document. That is the
same condition the No-Skill row of \btab{tab:main} reports, so AWM induces from
the same experience every arm starts out from, and from nothing our method
produced.

Retrieval follows the official implementation and is global. Every test task
queries the index, but the results are pooled rather than kept per task, and
the top-$k$ are written into one file that is then injected whole and
identically into every task. Embedding follows the official
\idt{text-\ab embedding-\ab ada-\ab 002}. The one knob not inherited is top-$k$, set to 3
rather than the official default of 10. The change is in AWM's favour: the
induced libraries here hold roughly 4 to 30 entries per cell, so the official
value would have degenerated retrieval into whole-library injection and removed
the retrieval step the method depends on. At deployment the workflow document is injected as the
initial skill, with one epoch and a single batch covering the whole task
stream.

\begin{apptab}
\setlength{\tabcolsep}{3pt}
\begin{tabular}{@{}l l r r r r r@{}}
\toprule
& & & \multicolumn{2}{c}{\textbf{AWM hard}} & \multicolumn{2}{c}{\textbf{AWM soft}} \\
\cmidrule(lr){4-5}\cmidrule(lr){6-7}
\textbf{Model} & \textbf{Bench} & $n$ & value & $\Delta$ & value & $\Delta$ \\
\midrule
\multirow{4}{*}{MiniMax-M3}
 & ALFWorld & 42 & 85.7 & \gainval{9.5} & 93.7 & \gainval{3.2} \\
 & TBP      & 32 & 34.4 & \dropval{6.2} & 75.1 & \dropval{2.0} \\
 & SWE      & 20 & 30.0 & \dropval{5.0} & 34.9 & \dropval{4.9} \\
 & LMB      & 53 & 24.5 & \gainval{1.9} & 24.5 & \gainval{1.9} \\
\midrule
\multirow{4}{*}{DeepSeek-V4-Pro}
 & ALFWorld & 42 & 76.2 & \gainval{\textbf{19.1}} & 85.7 & \gainval{17.8} \\
 & TBP      & 32 & 37.5 & \gainval{3.1}  & 70.4 & \gainval{1.5} \\
 & SWE      & 20 & 65.0 & \gainval{\textbf{20.0}} & 67.5 & \gainval{20.0} \\
 & LMB      & 53 & 13.2 & \naval         & 13.2 & \naval \\
\midrule
\multirow{4}{*}{GPT-5.4-mini}
 & ALFWorld & 42 & 45.2 & \gainval{2.3}  & 61.1 & \gainval{0.2} \\
 & TBP      & 32 & 40.6 & \gainval{12.5} & 78.3 & \gainval{2.7} \\
 & SWE      & 20 & 40.0 & \gainval{5.0}  & 45.8 & \gainval{8.4} \\
 & LMB      & 53 & 13.2 & \dropval{1.9}  & 13.2 & \dropval{1.9} \\
\midrule
\textbf{Mean} & & & \textbf{42.1} & \gainval{5.0} & & \gainval{3.9} \\
\bottomrule
\end{tabular}
\end{apptab}
\apptabcap{AWM across all twelve cells. $\Delta$ is against No-Skill in the
same cell of \btab{tab:main}, on the same scale as \btab{tab:ablation} ---
median \gainval{2.7} and \gainval{1.7}, eight up, one flat, three down. As
there, the arrow and not the colour carries the sign, and a cell with no
$\Delta$ did not move.}{tab:c-awm}

Table~\ref{tab:c-awm} gives every cell. Across the twelve cells AWM averages $+5.0$ with a per-cell range from $-6.2$
to $+20.0$. GLoW's Global arm on the same twelve cells is 12/12 same-signed
with a mean of $+17.2$.

Three differences correspond to our mechanism. AWM induces over the task set as
a whole without first forming families. That shows up most clearly on LMB, the
benchmark with the fewest mergeable tasks: the three models give $+1.9$, $0.0$
and $-1.9$, a mean of zero, against $+24.5$, $+5.7$ and $+3.8$ for GLoW's
Global arm in the same cells. Its induced workflows enter reuse with no admission step grounded
in real execution, whereas the margins of the candidates rejected in
\app{app:gate} lie between $-0.05$ and $-0.13$. And after retrieval it injects
one identical document into every task, whereas GLoW's Recall is per task.

The body compares ``one round of family consolidation'' against AWM's $+5.0$;
that figure is the round-0 committed library's real deployment score in the
first sub-round of round 1, minus the same cell's No-Skill, on hard, averaged
over the twelve cells.

\section{Statistical Tests}
\label{app:stats}

\subsection{Paired Tests Across Runs}
\label{app:paired}

The twelve cells of the main table are paired measurements on the same tasks
under the same harness. The per-cell differences of the Global arm, in
percentage points:

\begin{apptab}
\begin{tabular}{@{}l rrrr@{}}
\toprule
\textbf{Model} & \textbf{TBP} & \textbf{SWE} & \textbf{ALFWorld} & \textbf{LMB} \\
\midrule
DeepSeek-V4-Pro & $+15.6$ & $+25.0$ & $+26.2$ & $+24.5$ \\
MiniMax-M3      & $+12.5$ & $+25.0$ & $+16.7$ & $\phantom{0}+5.7$ \\
GPT-5.4-mini    & $+21.9$ & $+10.0$ & $+19.0$ & $\phantom{0}+3.8$ \\
\bottomrule
\end{tabular}
\end{apptab}
\apptabcap{Per-cell difference of the Global arm against No-Skill (hard,
pp).}{tab:d-deltas}

\begin{apptab}
\begin{tabular}{@{}l l r@{}}
\toprule
\textbf{Test} & \textbf{Statistic} & \textbf{Two-sided $p$} \\
\midrule
Sign test            & 12/12 positive  & \textbf{0.000488} \\
Wilcoxon signed-rank & $W_-=0$, $n=12$ & \textbf{0.000488} \\
\bottomrule
\end{tabular}
\end{apptab}
\apptabcap{Paired tests over the twelve runs.}{tab:d-tests}

The two paired tests are Table~\ref{tab:d-tests}. The Global$+$Local arm is
likewise 12/12 positive, with a mean of $+18.0$\,pp
and the same statistics. When every difference has the same sign the two tests
collapse to $p = 2/2^{12}$, the smallest $p$ attainable at $n=12$. The
significance is established at the level of the twelve runs; per-cell resolution is in
\app{app:percell}.

\subsection{What the Peak Read-out Is Taken Over}
\label{app:reading}

The main table reports a peak across rounds, and the range that peak is taken
over is restricted. GLoW's best-across-rounds is taken only among libraries
that passed the commit gate and were actually deployed, never among candidate
versions: a candidate that scores higher but is rejected by the gate does not
enter the pool. The solid line of \bfig{fig:gate} is that deployed trajectory
and the dashed line is the same round's candidate; in the last round they are
$+14.7$\,pp against $+9.6$\,pp, so the retained older library is 5.1\,pp above
the new candidate.

The compared side is on the same scale. The SkillOpt row takes the
epoch-weighted merged train peak; like GLoW's peak across rounds, it is a
maximum along its own optimisation axis.

\subsection{Gate Decisions and the Tolerance}
\label{app:gate}

$D$ in \beq{eq:gate} is the training task stream itself; held-out evaluation
enters no gate decision. Both terms of the anchor
$A=\max(V^{\star},V^{\mathrm{ns}})$ can be read off the supplementary
material: $V^{\star}$ is the real deployment score of the last committed round,
recorded in each run's library state, and $V^{\mathrm{ns}}$ is that cell's
no-skill baseline, kept in a file of its own. In the first round there is no
committed round yet, so the anchor reduces to the baseline.

Candidates are adjudicated late: the candidate of round $k$ is decided by the
first real deployment of round $k{+}1$. The direct consequence of that
discipline is that every prior in the library carries the evidence of one real
deployment, and a candidate that was never adjudicated enters none of the
libraries reported in this paper.

\textbf{$\epsilon=0.02$ did change outcomes.} Two candidates scored below the
anchor but fell inside the tolerance band, with margins of $-0.0108$ and
$-0.0181$ on the soft scale --- that is, $-1.08$ and $-1.81$ percentage points
--- both on TBP, and both were admitted; at $\epsilon=0$ both would have been
rejected. Both are smaller than the 2.71\,pp reproduction standard deviation
of repeated measurement (\app{app:dispersion}), so what the
tolerance absorbs is a dip of the size of the noise rather than a real
degradation --- which is the setting \bsec{sec:commit} describes as
``$\epsilon$ absorbs fluctuation''. The rejected candidates' margins lie
between $-0.05$ and $-0.13$, i.e.\ $-5$ to $-13$ percentage points, far outside
the tolerance of $\epsilon=0.02$ (2\,pp), and $\epsilon=0$ would not change
those decisions.

The gate includes equality, so a tie passes. That is a design choice, not a
conclusion we verify. The reason for admitting a tie is that once a
prior is in the library the next round solves on top of it and local skills are
extracted from those solutions, so a candidate that merely holds level can
still change what the following round produces. Whether that helps or hurts on
balance was not tested.

\section{Library-Size Accounting}
\label{app:library-size}

The compression ratio compares that run's last-round committed global priors
against the pool of local skills, one per task, from the same round. The entry ratio
counts entries; the word ratio sums word counts on both sides. The denominator
takes only the final state of the local skills, not the per-sub-round
intermediate snapshots: on ALFWorld, counting those too would give
$42+3\times42=168$ entries in place of 42, inflating the denominator fourfold.

\begin{apptabsm}
\setlength{\tabcolsep}{2pt}
\begin{tabular}{@{} l l rrr r rr r@{}}
\toprule
& & \multicolumn{3}{c}{\textbf{Entries}} & &
    \multicolumn{2}{c}{\textbf{Words}} & \\
\cmidrule(lr){3-5}\cmidrule(lr){7-8}
\textbf{Benchmark} & \textbf{Model}
 & \hdr{priors}{}& \hdr{local}{} & \hdr{tasks}{expected}
 & \hdr{entry}{ratio}
 & \hdr{prior}{} & \hdr{local}{}
 & \hdr{word}{ratio} \\
\midrule
\multirow{3}{*}{TBP}
 & DeepSeek-V4-Pro & 16 & 31 & 32 & 0.516 & 11955 & 32335 & 0.370 \\
 & MiniMax-M3      & 10 & 32 & 32 & 0.312 & \phantom{0}8700 & 53778 & 0.162 \\
 & GPT-5.4-mini    & 11 & 32 & 32 & 0.344 & \phantom{0}8067 & 36089 & 0.224 \\
\midrule
\multirow{3}{*}{SWE}
 & DeepSeek-V4-Pro & \phantom{0}8 & 20 & 20 & 0.400 & \phantom{0}5741 & 16251 & 0.353 \\
 & MiniMax-M3      & \phantom{0}8 & 20 & 20 & 0.400 & \phantom{0}7589 & 21658 & 0.350 \\
 & GPT-5.4-mini    & \phantom{0}6 & 20 & 20 & 0.300 & \phantom{0}4434 & \phantom{0}9900 & 0.448 \\
\midrule
\multirow{3}{*}{ALFWorld}
 & DeepSeek-V4-Pro & 15 & 42 & 42 & 0.357 & 10876 & 36212 & 0.300 \\
 & MiniMax-M3      & 16 & 42 & 42 & 0.381 & 15204 & 57912 & 0.263 \\
 & GPT-5.4-mini    & 12 & 42 & 42 & 0.286 & \phantom{0}7801 & 36059 & 0.216 \\
\midrule
\multirow{3}{*}{LMB}
 & DeepSeek-V4-Pro & 13 & 53 & 53 & 0.245 & \phantom{0}9123 & 28546 & 0.320 \\
 & MiniMax-M3      & \phantom{0}6 & 53 & 53 & 0.113 & \phantom{0}5104 & 25708 & 0.199 \\
 & GPT-5.4-mini    & \phantom{0}3 & 53 & 53 & 0.057 & \phantom{00}692 & \phantom{0}5266 & 0.131 \\
\bottomrule
\end{tabular}
\end{apptabsm}
\apptabcap{All twelve cells at the last round. Lower is more compression.}{tab:e-percell}

\begin{apptabsm}
\begin{tabular}{@{}l rr rr@{}}
\toprule
\textbf{All twelve cells} & \multicolumn{2}{c}{\textbf{Entries}}
 & \multicolumn{2}{c}{\textbf{Words}} \\
\cmidrule(lr){2-3}\cmidrule(lr){4-5}
& ratio & $1/$ratio & ratio & $1/$ratio \\
\midrule
round 1                & 0.299$\times$ & 3.35$\times$ & 0.278$\times$ & 3.60$\times$ \\
\textbf{round 2}, last & \textbf{0.309$\times$} & \textbf{3.23$\times$}
                       & \textbf{0.278$\times$} & \textbf{3.60$\times$} \\
\bottomrule
\end{tabular}
\end{apptabsm}
\apptabcap{Summary over both rounds and both read-outs.}{tab:e-summary}

Table~\ref{tab:e-percell} gives every cell and Table~\ref{tab:e-summary} the
summary; all four combinations of read-out land between $0.28$ and $0.31$. The body
takes the reciprocal of the last round's word ratio and states it uniformly as
$3.6\times$ compression. Of the three aggregations available, the body reports
the one least favourable to us: pooling the word counts on both sides before
dividing gives $3.8\times$, and the mean of the per-cell reciprocals gives
$4.1\times$.

\section{Held-Out Transfer Results}
\label{app:transfer}

\subsection{ALFWorld Held-Out}
\label{app:transfer-alf}

\bsec{sec:main}, ``On unseen tasks the prior alone carries the gain'', refers
to the Global column of \btab{tab:transfer}: the frozen consolidated prior
injected on its own, with no local regeneration. The three rows come from the
runs below, all on the 60-task official unseen split at a single trial.

\begin{apptabsm}
\begin{tabular}{@{}l rr p{0.40\columnwidth}@{}}
\toprule
\textbf{Model} & \textbf{No-Skill} & \textbf{Global} & \textbf{Runs} \\
\midrule
DeepSeek-V4-Pro & 71.7 & \textbf{88.3}
 & \idt{heldout\_alf60\_ds\_t1\_noskill\_0722} /\ \idt{\dots\_t1\_global\_0722} \\
GPT-5.4-mini    & 66.7 & \textbf{75.0}
 & \idt{heldout\_alf60\_gpt\_t1\_noskill\_0722} /\ \idt{\dots\_globalR1\_0723} \\
MiniMax-M3      & 83.3 & \textbf{88.3}
 & \idt{heldout\_alf60\_m3\_t1\_noskill\_0722b} /\ \idt{\dots\_t1\_global\_0722b} \\
\midrule
Mean            & 73.9 & \textbf{83.9} & \\
\bottomrule
\end{tabular}
\end{apptabsm}
\apptabcap{ALFWorld held-out, 60 unseen tasks, hard (\%).}{tab:f-alfworld}

All three models move the same way, and MiniMax-M3, which starts highest, gains
least --- consistent with a ceiling effect. The library received no
modification of any kind for the unseen tasks: what is injected is the version
committed on the training stream.

\subsection{SWE Held-Out}
\label{app:transfer-swe}

The body's $40.0 \rightarrow 45.6$ comes from the matched pair of three-trial
runs named in the table below, each reported as the mean resolve rate over its
three trials. The benchmark is
SWE's 30-task held-out split, disjoint from train20; the model is MiniMax-M3.

\begin{apptabsm}
\begin{tabular}{@{}p{0.40\columnwidth} r r r@{}}
\toprule
\textbf{Run} & \textbf{trials} & \textbf{priors} & \textbf{resolve rate} \\
\midrule
\idt{heldout\_swe30\_noskill\_m3\_0709}        & 3 & 0 & \textbf{0.4000} \\
\idt{heldout\_swe30\_global\_round0\_m3\_0709} & 3 & 5 & \textbf{0.4556} \\
\bottomrule
\end{tabular}
\end{apptabsm}
\apptabcap{The pair of runs behind the $+5.6$\,pp.}{tab:f-swe}

\section{What the Committed Priors Contain}
\label{app:skills}

\subsection{Agreement With the Dataset's Labels}
\label{app:alf-labels}

Of the four benchmarks only ALFWorld carries solution-type labels independent
of our algorithm, seven tasks in each of six types.

\begin{apptabsm}
\begin{tabular}{@{}l r r r r r r@{}}
\toprule
\textbf{Round} & $K$ & \textbf{purity} & \textbf{ARI} & \textbf{NMI}
 & \hdr{single-type}{families} & \hdr{clustered}{tasks} \\
\midrule
round 2 & 15 & \textbf{0.950} & 0.605 & 0.801 & 13\,/\,15 & 40 \\
\bottomrule
\end{tabular}
\end{apptabsm}
\apptabcap{Agreement between our families and ALFWorld's six canonical task
types.}{tab:g-alf-agreement}

Table~\ref{tab:g-alf-agreement} gives the agreement scores. Purity is high while
ARI is middling, and the gap comes from over-segmentation
rather than from mixing types. $K=15$ against 6 types, and ARI penalises
splitting one type, and splitting is exactly the design intent: families are
finer-grained than task types. The high purity is not bought with fragmentation
either --- 5 of the 15 families are singletons, and removing them leaves the
remaining 35 tasks still at a purity of 0.943.

Two families cut across the dataset's goal taxonomy along a shared procedure.
The quotations are the procedure summaries written by the compression stage.

\begin{apptabsm}
\begin{tabular}{@{} p{0.20\columnwidth} p{0.13\columnwidth} p{0.35\columnwidth} p{0.17\columnwidth}@{}}
\toprule
\textbf{Cross-cutting family} & \textbf{Member type}
 & \textbf{Procedure summary} & \textbf{Shared procedure} \\
\midrule
\idt{household\_pick-\ab and-\ab place:\ab \_locate}
 & \idt{pick\_and\_place}
 & ``locate egg on surfaces, \textbf{open fridge, move object to interior
    compartment}''
 & place into a closed container \\
 & \idt{pick\_two}
 & ``Glassbottle placement into \textbf{closed fridge via single-carry
    open-before-place gate}'' & \\
\midrule
\idt{household\_object\_pick-\ab cool-\ab place}
 & \idt{pick\_cool}
 & ``locate item, \textbf{cool via appliance, deposit} at target receptacle''
 & deposit after an appliance transform \\
 & \idt{pick\_heat}
 & ``locate, pick up, \textbf{heat via appliance, place} in target receptacle''
 & \\
\bottomrule
\end{tabular}
\end{apptabsm}
\apptabcap{The dataset's six types are divided by goal, and both cross-cuts fall
where the goals differ and the procedure does not. The second is the clearer
case: both tasks are an Egg, one cooled and then placed in the microwave, the
other heated and then placed in the fridge --- appliance and receptacle
swapped, procedural skeleton identical.}{tab:g-crosscut}

\subsection{The Family Mapping for the Terminal Benchmark}
\label{app:terminal-map}

The mapping \bsec{sec:analysis} points at: TBP\,$\times$\,GPT-5.4-mini, round 2
against the round-0 baseline, hard, paired task by task.

\begin{apptabsm}
\begin{tabular}{@{} p{0.32\columnwidth} r c r r@{}}
\toprule
\textbf{Family} & $n$ & \textbf{base\,$\to$\,r2} & \textbf{fixed} & \textbf{broken} \\
\midrule
\idt{postgresql\_recovery\_sanitization} & 5 & 1\,$\to$\,3 & \gainval{2} & \naval \\
\idt{flooded-\ab grid\_escape\_planning}     & 5 & 4\,$\to$\,4 & \gainval{1} & \dropval{1} \\
\idt{stripped\_binary\_feature-\ab flag}     & 5 & 0\,$\to$\,2 & \gainval{2} & \naval \\
\idt{two-\ab dimensional\_gaussian\_mixture} & 4 & 2\,$\to$\,2 & \naval & \naval \\
\idt{csv-\ab to-\ab json\_organizational\_merge} & 3 & 0\,$\to$\,1 & \gainval{1} & \naval \\
\idt{latex\_document\_build}             & 3 & 1\,$\to$\,3 & \gainval{2} & \naval \\
\idt{shellcode\_network\_tracing}        & 2 & 1\,$\to$\,1 & \naval & \naval \\
\idt{user\_profile\_grpc}                & 2 & 0\,$\to$\,0 & \naval & \naval \\
\idt{apache\_virtual\_host}              & 2 & 0\,$\to$\,1 & \gainval{1} & \naval \\
\idt{distributed\_tensor-\ab parallel\_matrix} & 1 & 0\,$\to$\,0 & \naval & \naval \\
\midrule
\textbf{Ten families} & \textbf{32} & \textbf{9\,$\to$\,17}
 & \gainval{\textbf{9}} & \dropval{\textbf{1}} \\
\bottomrule
\end{tabular}
\end{apptabsm}
\apptabcap{The full terminal mapping: ten families over 32 tasks, nine repairs
against one break --- \bsec{sec:analysis}, ``9 repairs against one
pass-to-fail''. A dash means no task moved in that direction.}{tab:g-terminal}

The mapping is Table~\ref{tab:g-terminal}, and the split inside one family is
Table~\ref{tab:g-stripped}. The match happens at the procedure level: all five tasks in
\idt{stripped\_binary\_feature-\ab flag} are about binaries by subject, but they
split on whether the solving procedure really operates at the byte level. That
family's prior opens with ``Scan repeated markers'', followed by validating a
candidate manifest and decrypting bitmaps.

\begin{apptabsm}
\begin{tabular}{@{}p{0.22\columnwidth} p{0.30\columnwidth} c c c@{}}
\toprule
\textbf{Member task} & \textbf{Actual solving procedure}
 & \hdr{matches the}{family prior}
 & \textbf{no-skill} & \textbf{round 2} \\
\midrule
\idt{detect-\ab c-\ab feature-\ab flags}
 & reverse the ELF, validate the manifest, decrypt bitmaps & $\checkmark$ & 0 & \textbf{PASS} \\
\idt{implement-\ab lz77-\ab file-\ab compressor}
 & \textbf{scan raw bytes} for greedy longest-match packing & $\checkmark$ & 0 & \textbf{PASS} \\
\idt{implement-\ab crc32-\ab with-\ab logic-\ab gates}
 & synthesise a logic-gate netlist & & 0 & fail \\
\idt{recover-\ab stream-\ab cipher-\ab key}
 & brute-force the keystream mathematically & & 0 & fail \\
\idt{regex-\ab bitcoin-\ab p2pkh-\ab extraction}
 & construct a regular expression & & 0 & fail \\
\bottomrule
\end{tabular}
\end{apptabsm}
\apptabcap{The prior's procedure is scanning for markers and parsing fields, and
it repaired only the two tasks that really do work at the byte level. The test
does not depend on clustering quality: both input columns are facts external to
the algorithm --- each task's real solution is read from the trajectory, and
whether it passed is decided by the verifier.}{tab:g-stripped}

\subsection{Excerpts of Prior Text}
\label{app:excerpts}

Three excerpts, ordered by how much the family actually shares, from least to
most.

The least shared comes from a DeepSeek run on a 16-task TBP split whose
families were formed by pre-partitioning the tasks by domain; that run is
\emph{not} one of the twelve in the main table. Its debugging family prior runs
to 500 words and describes itself as

\begin{priorquote}
Establish baseline execution, gather direct runtime evidence, and report
conclusions conservatively.
\end{priorquote}

\noindent
The merge note the compressor wrote for itself is ``\emph{These 2 tasks span
unrelated domains and \textbf{share only generic engineering discipline}}''.
The notes of the same run's security and machine-learning family priors
likewise record that no shared algorithm was found; the three are 500, 426 and
384 words. This is direct evidence for \bsec{sec:ablation}, ``a merged
document yields generic discipline the model has already internalized''.

A finer family comes from round 2 of TBP\,$\times$\,GPT-5.4-mini --- the prior
of the \idt{stripped\_binary\_feature-\ab flag} family of \app{app:terminal-map},
650 words.

\begin{priorquote}
Scan repeated markers, validate candidate manifests, decrypt feature bitmaps,
and emit ordered flags.
\end{priorquote}

\noindent
This is also a terminal task, but because the family's members genuinely share
byte-level work, compression kept domain-level steps such as manifest
validation and bitmap decryption. The remaining priors of the same run carry
domain shape too: recovering a PostgreSQL key from the WAL (833 words), step
selection for flood-grid hazard avoidance (786), deterministic EM fitting of a
two-dimensional Gaussian mixture (814), and building a deterministic LaTeX PDF
in an isolated temporary directory (777).

The most shared comes from a round-1 family prior of
ALFWorld\,$\times$\,MiniMax-M3, 926 words in all, of which its procedure
section is 440.

\begin{priorquote}
Step 1 $\mid$ \textbf{Action}: Scan the initial observation for the
illuminator-class object (lamp\,/\,mirror\,/\,magnifier\,/\,lantern) and record
its receptacle as the fixed final destination $\mid$ \textbf{Pitfall}:
\textbf{Do not default to `desk'} --- illuminators frequently sit on dressers,
nightstands, or shelves.
\\[3pt]
Step 2 $\mid$ \textbf{Action}: Issue \idt{go to \textless source\_receptacle\textgreater} then
\idt{take \textless target\textgreater\ from \textless source\_receptacle\textgreater} directly $\mid$ \textbf{Pitfall}:
A separate \idt{examine} wastes a turn\dots
\end{priorquote}

\noindent
\bsec{sec:analysis}, ``What a prior keeps'', corresponds to this one. One of
the seven local skills in that family uses 989 words for seven steps and embeds
a memorised location table; the compressed family prior states the same job in
four steps and 440 words of procedure, the table is gone, and Step~1 becomes a
single read. The difference from
the first excerpt is not length but executability: this one carries concrete
action commands and a turn-budget consideration and can change the search path,
while the first can only change the wording. Its merge note also records that
the compressor chose between two mutually exclusive strategies rather than
concatenating them.

\section{Benchmarks and Splits}
\label{app:splits}

There is one set of splits in this paper, and everything reads it. All four
task streams are laid out once and held fixed across rounds; the
twelve runs and both compared methods read the same split files, and no cell
uses a task set drawn separately for it. All four split files record
\idt{seed=42}, and that seed controls the TBP, SWE and LMB
splits; the ALFWorld split is a deterministic selection and does not depend on
it. The split files ship with the supplementary
material, each recording its split name, seed and size; the LMB and ALFWorld
files additionally carry a hash of the task identifiers.

\begin{apptab}
\begin{tabular}{@{} p{0.27\columnwidth} p{0.21\columnwidth} p{0.39\columnwidth}@{}}
\toprule
\textbf{Benchmark} & \textbf{Stream size} & \textbf{Basis of the split} \\
\midrule
Terminal-Bench-Pro
  & 32 tasks, \idt{tbp\_hard\_train32}
  & the hard subset, 4 tasks from each of 8 classes \\
SWE-bench Verified
  & 20 instances, \idt{swe\_15min1h\_train20}
  & the official Verified set, filtered by the official difficulty annotation
    to the 15 min -- 1 hour band \\
ALFWorld
  & 42 tasks, \idt{alfworld\_train42}
  & derived from \idt{alfworld\_train60}, taking the lexicographically first 7
    of each task type --- deterministic, not random sampling \\
Live\-Mathematician\-Bench
  & 53 tasks, \idt{lmb\_train53}
  & the official HuggingFace monthly files 202511--202602, merging the upstream
    train (35) and val (18) \\
\bottomrule
\end{tabular}
\end{apptab}
\apptabcap{The four task streams and how each was cut.}{tab:h-splits}

The four streams are Table~\ref{tab:h-splits}. This paper has exactly two
held-out evaluations, listed in Table~\ref{tab:h-heldout}, and they come from
different places.

\begin{apptab}
\begin{tabular}{@{}p{0.26\columnwidth} p{0.28\columnwidth} p{0.36\columnwidth}@{}}
\toprule
\textbf{Used for} & \textbf{Set} & \textbf{Origin} \\
\midrule
Transfer, \btab{tab:transfer}
 & ALFWorld \idt{valid\_unseen}, 60 tasks
 & the dataset's own official unseen split \\
\bsec{sec:transfer}, software repair
 & \idt{swe\_15min1h\_test30}, 30 instances
 & cut here, same source and same seed as train20 \\
\bottomrule
\end{tabular}
\end{apptab}
\apptabcap{The two held-out sets. They differ in origin, so they carry
different kinds of evidential weight.}{tab:h-heldout}

\idt{seed=42} governs the sampled splits and the task traversal order, not
generation.
The base models are called through vendor APIs and decode at the default
temperature, measured at 0.7, and the seed does not enter the LLM request.

\section{Run Artefacts and Execution Volume}
\label{app:runs}

Every cell of \btab{tab:main} corresponds to one independent continual run. The
outputs of all twelve ship with the supplementary material, one directory per
cell, holding the per-round summaries, the complete configuration, the library
state, the per-task results, and both the committed priors and the per-task
local skills.

The twelve runs share one loop size: 3 rounds $\times$ 3 sub-rounds, one trial
per task. A sub-round is one real deployment over the whole task stream. The
loop size is the same for every benchmark and every model; no cell was given a
longer run than another.

\begin{apptab}
\begin{tabular}{@{}l r r r@{}}
\toprule
\textbf{Benchmark} & \textbf{Tasks}
 & \hdr{deployments}{per run} & \hdr{all three}{models} \\
\midrule
Terminal-Bench-Pro     & 32 & 288 & \phantom{0}864 \\
SWE-bench Verified     & 20 & 180 & \phantom{0}540 \\
ALFWorld               & 42 & 378 & 1134 \\
Live\-Mathematician\-Bench & 53 & 477 & 1431 \\
\midrule
\textbf{Total}         &    &     & \textbf{3969} \\
\bottomrule
\end{tabular}
\end{apptab}
\apptabcap{Training deployments.}{tab:i-budget}

The deployment count is Table~\ref{tab:i-budget}. Searching over the three
candidate routes buys no extra rollouts: the routes are not scored by running
the task stream again for
each; their scores are read off the nine deployments already executed, and
whether a candidate enters the library is settled by the first deployment of
the next round, which is already inside the budget. The table above is
therefore the entire training cost, and none of the machinery this paper adds
is paid for outside it.

Sub-rounds have diminishing returns. Pooling the twelve runs, three rounds and
all tasks gives 1323 task-$\times$-round pairs. The second sub-round genuinely
improves a task's best local skill 162 times, 12.2\%, and the third 70
times, 5.3\%.

\section{Reproduction Dispersion and Resolution}
\label{app:dispersion-section}

\subsection{Reproduction Dispersion of a Single Trial}
\label{app:dispersion}

Decoding is not deterministic, so the reproduction spread is measured rather
than assumed. On the four splits the main table uses, we take every round-0
no-injection measurement matching on split, model, trial count and time limit
--- 18 measurements in six groups.

\begin{apptabsm}
\begin{tabular}{@{}p{0.36\columnwidth} r r r r@{}}
\toprule
\textbf{Split $\times$ Model} & \textbf{reps}
 & \textbf{mean} & \textbf{s.d.} & \textbf{range} \\
\midrule
SWE \idt{train20} $\times$ DeepSeek-V4-Pro      & 3 & 45.00\% & \textbf{0.00} & 0.0\,pp \\
SWE \idt{train20} $\times$ MiniMax-M3           & 2 & 37.50\% & 3.54 & 5.0\,pp \\
ALFWorld \idt{train42} $\times$ MiniMax-M3      & 2 & 73.81\% & 3.37 & 4.8\,pp \\
ALFWorld \idt{train42} $\times$ DeepSeek-V4-Pro & 2 & 59.52\% & 3.37 & 4.8\,pp \\
LMB \idt{train53} $\times$ MiniMax-M3           & 3 & 24.53\% & 3.27 & 5.7\,pp \\
LMB \idt{train53} $\times$ GPT-5.4-mini         & 6 & 14.78\% & 2.51 & 5.7\,pp \\
\bottomrule
\end{tabular}
\end{apptabsm}
\apptabcap{Repeated round-0 No-Skill measurements.}{tab:j-dispersion}

Table~\ref{tab:j-dispersion} lists the six groups. The pooled within-group
standard deviation is 2.71\,pp on 12 degrees of
freedom. The main table's mean Global gain over no-skill is 17.2 points, 6.4
times that standard deviation, and all twelve runs share a sign.

\subsection{Per-Cell Resolution}
\label{app:percell}

The significance in \app{app:paired} belongs to the level of the twelve runs
and cannot be pushed down to a single cell. Single-cell resolution is taken
from the measured value in \app{app:dispersion}: with a reproduction standard
deviation of 2.71\,pp, two standard deviations is $\pm 5.4$\,pp, and a reading
inside that range is within reproduction spread. That scale comes from 18 real
repetitions and does not rest on an assumption that tasks are independent and
identically distributed. Under that assumption, 42 and 53 tasks would give
$\pm 9$ to $15$\,pp, two to three times wider than what was measured.

\section{Hyperparameters}
\label{app:hparams}

There is one hyperparameter setting in this paper, not twelve. Every entry in
Table~\ref{tab:k-hparams} takes one value across all four
benchmarks $\times$ three models, and not one of them was adjusted for a cell
by that cell's own result. The gains in the twelve cells are therefore not the
product of per-cell search; the price is that this paper does not report a
sensitivity analysis for any of these values.

\begin{apptabsm}
\begin{tabular}{@{} p{0.20\columnwidth} p{0.32\columnwidth} p{0.40\columnwidth}@{}}
\toprule
\textbf{Parameter} & \textbf{Value} & \textbf{Selection criterion} \\
\midrule
Rounds $T$ & 3
 & performance peaks at round 1 and then decays; three rounds are enough to
   expose that \\
Sub-rounds $J$ & 3 & marginal returns in \app{app:runs} \\
Trials per task & 1
 & cost; dispersion in \app{app:dispersion}, resolution in \app{app:percell} \\
Clustering view weights
 & five-view 0.30/\ab 0.20/\ab 0.20/\ab 0.20/\ab 0.10;
   four-view $0.25\times4$
 & hard-coded constants; provenance in \app{app:fusion} \\
Number of clusters $K$ & Kneedle knee, chosen automatically
 & not set by hand; see \app{app:fusion} \\
Recall top-$k$ & 1
 & injected only above threshold, otherwise the base prior alone \\
Recall similarity threshold & 0.45 & a single value shared by all four benchmarks \\
Commit-gate tolerance $\epsilon$ & 0.02
 & smaller than the measured reproduction standard deviation, so that it
   absorbs noise; see \app{app:gate} \\
Embedding quantisation & 4-bit
 & 8-bit was measured not to change routing decisions \\
Decoding temperature & vendor default, measured 0.7 & unchanged \\
\bottomrule
\end{tabular}
\end{apptabsm}
\apptabcap{The complete hyperparameter table.}{tab:k-hparams}

\section{Compute Environment}
\label{app:environment}

\begin{apptab}
\begin{tabular}{@{}l p{0.62\columnwidth}@{}}
\toprule
\textbf{Item} & \textbf{Value} \\
\midrule
CPU              & AMD Ryzen AI 9 365, 18 logical cores \\
GPU              & NVIDIA RTX 5080 Laptop, 16\,GB VRAM, driver 610.74 \\
Memory           & 21\,GB allocated to WSL2, host 31\,GB \\
Operating system & Windows $+$ WSL2, Linux kernel 6.18.33.2 \\
Container        & Docker 29.3.0, for the SWE-bench and Terminal-Bench-Pro
                   task containers \\
Python           & 3.13.12, conda environment \\
Key libraries    & torch 2.11.0 $\cdot$ transformers 5.8.0 $\cdot$
                   scikit-learn 1.8.0 $\cdot$ bitsandbytes 0.49.2 \\
Embedding model  & Qwen3-Embedding-8B, 4-bit quantisation, about 5.8\,GB
                   resident VRAM \\
Base models      & DeepSeek-V4-Pro, MiniMax-M3, GPT-5.4-mini, all through
                   vendor APIs \\
\bottomrule
\end{tabular}
\end{apptab}
\apptabcap{Compute environment.}{tab:l-environment}

The environment is Table~\ref{tab:l-environment}. The WSL2 memory ceiling is a
measured constraint rather than a preference: at
24\,GB a fully loaded SWE run exhausts host memory, so it was lowered to
21\,GB, and the concurrency of SWE and TBP is bounded by that. The GPU is used
only for embedding retrieval and takes no part in base-model inference.

\section{Consolidation-Pipeline Prompts}
\label{app:prompts}

The consolidation pipeline uses 24 prompt templates in total, and every one of
them is phrased benchmark-independently: no benchmark name and no task
identifier, and the only domain-specific text anywhere in them is a single
category phrase, described at the end of this section. The five core prompts below are given in pipeline order and
are the variants actually deployed in the twelve runs; the placeholders are
filled with card content at run time.

They are set full width so that they keep the exact line breaking of the
source; nothing has been re-wrapped or abridged.

\begin{figure*}[t]
\begin{promptcard}{Local skill extraction}{written into the procedure section after a task turns from failing to passing; de-instantiation is enforced at the top of the pipeline}
\begin{verbatim}
A task just succeeded (pass_rate=1.0) after previously failing (pass_rate=0.0).
Extract MINIMAL, reusable procedure knowledge for the skill's procedure slot.
Successful trajectory (key steps): {traj}
Final code (key part): {code}
Rules:
- Output 2-4 concise bullet points describing the STEPS that made this work
- Focus on FLOW: what to do first, what to check, what to call
- Include at most ONE short code snippet (<=8 lines) only if it encodes a
  non-obvious pattern
- Do NOT include task-specific file paths, test constants, or verifier output
- Each bullet must be actionable: start with a verb
- Total output: <=80 words
\end{verbatim}
\end{promptcard}

\begin{promptcard}{Failure attribution}{drives the append-only repair route}
\begin{verbatim}
An AI agent attempted a task but failed. Your job: identify which skill slot's
guidance was missing or wrong, based on the agent's actual code.
Extract the missing reusable knowledge, rule, mapping, schema invariant, parser
assumption, API/library usage rule, numeric/statistical criterion, or
algorithmic constraint that would have prevented this failure.
Do NOT default to generic process advice such as "avoid loops", "stop earlier",
or "be more careful" unless the evidence clearly proves that is the primary
missing rule.
\end{verbatim}
\end{promptcard}
\caption{The two prompts that read a single task's own execution.}
\label{fig:m-prompts-12}
\end{figure*}

\begin{figure*}[t]
\begin{promptcard}{Pairwise procedure judgement}{the LLM channel of clustering; this is how the body's ``groups by how it is solved rather than what the task is about'' is implemented}
\begin{verbatim}
You are given two task descriptions.
Task 1: {a}
Task 2: {b}
Decide whether they share the SAME reusable procedure: would one procedure,
written once, correctly accomplish the core of BOTH tasks? Judge by the
underlying operation and the shape of the data flow, not by surface wording
or subject matter.
Return ONLY JSON: {"same": true | false, "reason": "<short>"}
\end{verbatim}
\end{promptcard}

\begin{promptcard}{Canonical signature generation}{feeds the signature channels of multi-view fusion}
\begin{verbatim}
You are organizing {n} skill cards into a small set of canonical operation
categories. Group these {n} cards into K canonical operation buckets where K is
your choice in [{min_k}, {max_k}]. Two cards belong in the same bucket iff they
perform the SAME FUNDAMENTAL OPERATION (differing only in surface details:
column names, thresholds, file paths, domain). They MUST NOT be merged just
because they touch the same surface tech.
For each bucket, emit ONE canonical_label (3-5 lowercase words, lead with a
generic verb). All cards in a bucket get this EXACT canonical_label as their
new compact signature.
\end{verbatim}
\end{promptcard}
\caption{The two prompts that decide grouping.}
\label{fig:m-prompts-34}
\end{figure*}

\begin{figure*}[t]
\begin{promptcard}{Family compression}{gathers a family's local skills into one candidate global prior, i.e.\ \beq{eq:compress}}
\begin{verbatim}
You are distilling {n_tasks} task-specific skills into ONE reusable
family-level skill.
These tasks were grouped by an automated clustering step and may or may not
share a real solving procedure - do not assume they do. Before writing
anything, judge honestly: would solving one of these tasks actually teach you
concrete, reusable steps for solving the others (same algorithm family, same
class of checks, same kind of artifact), or do they only share generic
competent-engineering discipline (read the spec first, probe the environment,
enumerate edge cases, validate formally)? If a candidate Step/Pattern would
only ever fire for 1 of the {n_tasks} tasks, DO NOT include it - leave it out
rather than force it in. It is a CORRECT and USEFUL output for
section_procedure to end up short and generic if that is what genuinely
generalizes; a short honest skill is far more useful downstream than a long one
that silently overfits to whichever task you processed last.
\end{verbatim}
\end{promptcard}
\caption{The compression step. The compressor is explicitly permitted to return
something short and generic and is told that a forced step is harmful; the
first excerpt of \app{app:excerpts} is what this instruction produces.}
\label{fig:m-prompts-5}
\end{figure*}

None of the five contains a benchmark name, a task identifier or an answer. The
only domain interface is one placeholder, which at run time is replaced by the
current benchmark's category phrase --- \idt{terminal}, \idt{software
engineering}, \idt{household task} and \idt{research-\ab level math MCQ}
respectively. Family names are generated by the compression
stage itself and are not drawn from any preset taxonomy.

\section{Family Structure}
\label{app:families}

For each benchmark we take the run whose family structure is most evenly balanced, by the
measure below.
ALFWorld is selected by ARI against the dataset's labels; the other three have
no independent labels and are selected by the balance of the size distribution,
$\bigl[H(\text{sizes})/\ln K\bigr]\times(1-\text{singleton fraction})$. Prior counts
follow the read-out of \app{app:library-size} and include the one base prior
injected unconditionally into every library, so they are always the number of
families $K$ plus one; the \emph{Priors} column of \btab{tab:class-counts}
counts family priors only.

\subsection{Size and Distribution}
\label{app:family-structure}

\begin{apptabsm}
\setlength{\tabcolsep}{2pt}
\begin{tabular}{@{} p{0.26\columnwidth} r r r p{0.21\columnwidth} r r@{}}
\toprule
\textbf{Benchmark $\times$ Model} & \hdr{task}{stream} & \hdr{clus-}{tered}
 & $K$ & \textbf{Family sizes} & \hdr{committed}{priors} & \hdr{prior}{words} \\
\midrule
ALFWorld $\times$ MiniMax-M3 & 42 & 40 & 15
 & {\scriptsize 7,\ab 5,\ab 5,\ab 4,\ab 3,\ab 3,\ab 2,\ab 2,\ab 2,\ab 2,\ab 1$\times$5} & 16 & 15,204 \\
TBP $\times$ GPT-5.4-mini & 32 & 32 & 10
 & {\scriptsize 5,\ab 5,\ab 5,\ab 4,\ab 3,\ab 3,\ab 2,\ab 2,\ab 2,\ab 1} & 11 & \phantom{0}8,067 \\
SWE $\times$ GPT-5.4-mini & 20 & 11 & \phantom{0}5
 & {\scriptsize 4,\ab 2,\ab 2,\ab 2,\ab 1} & \phantom{0}6 & \phantom{0}4,434 \\
LMB $\times$ MiniMax-M3 & 53 & 14 & \phantom{0}5
 & {\scriptsize 5,\ab 4,\ab 3,\ab 1,\ab 1} & \phantom{0}6 & \phantom{0}5,104 \\
\bottomrule
\end{tabular}
\end{apptabsm}
\apptabcap{Family structure, one run per benchmark.}{tab:o-structure}

\subsection{Family Names and the Clustering Axis}
\label{app:family-names}

Family names are generated by the compression stage after clustering is
finished, and were not rewritten by hand. \textbf{The name is not the
clustering axis.} The namer takes the subject wording of some salient member of
the family, which is why SWE's family names carry a repository name and LMB's
carry the name of a branch of mathematics --- but the members are not grouped
along either of those axes.

\begin{apptabsm}
\begin{tabular}{@{} l p{0.34\columnwidth} p{0.44\columnwidth}@{}}
\toprule
\textbf{Benchmark} & \textbf{Family name} & \textbf{Member composition} \\
\midrule
\multirow{4}{*}{SWE}
 & \idt{django\_numeric\_expression}      & django $\times$3 $+$ \textbf{sympy} $\times$1 \\
 & \idt{astropy\_ascii\_html}             & astropy $\times$1 $+$ \textbf{django} $\times$1 \\
 & \idt{matplotlib\_image\_normalization} & matplotlib $\times$1 $+$ \textbf{xarray} $\times$1 \\
 & \idt{django\_queryset\_filtering}      & django $\times$2 \\
\midrule
\multirow{3}{*}{LMB}
 & \idt{asymptotically\_flat\_nonnegative}
 & manifold mass inequality, stochastic heat equation, dispersive PDE \\
 & \idt{exponential\_o-\ab minimal\_theory}
 & character varieties, ultrafilter orders, graph 1-factors, hyperbolic
   groups, transseries \\
 & \idt{primitive\_permutation\_group}
 & permutation groups, Chow quotients of flag varieties, Willmore surfaces,
   commutator length \\
\bottomrule
\end{tabular}
\end{apptabsm}
\apptabcap{Three of SWE's four multi-member families cross repositories, and all
three multi-member LMB families cross branches of mathematics.}{tab:o-names}

Table~\ref{tab:o-names} sets the names beside the member composition. What the
members really share sits in their procedure signatures. The three LMB
families' signatures are ``pick the strongest provable one among nested
propositions'', ``order by provable strength'' and ``give a complete
classification''; SWE's are ``locate the expression-handling table and add one
case'' and ``guard before passing downstream''. That is consistent with
\bsec{sec:cluster}, ``groups by how it is solved rather than what the task is
about''.

A reader who looks only at the family names would come away with the opposite
impression, which is why this section gives the member composition. The namer
sees only the text inside a family, never the relations between families, and
making the name reflect the clustering axis is not among its objectives.
\bsec{sec:analysis} records another instance of the same phenomenon on the
other TBP run: TBP\,$\times$\,MiniMax-M3 forms $K=9$ families, one of which
holds eight tasks --- a debugger session, WAL recovery and a logic-gate CRC32
among them --- and is named \idt{segfault\_debugging\_with} after one member.
That is a different partition from the TBP\,$\times$\,GPT-5.4-mini run of
\app{app:terminal-map}, which forms $K=10$ families with at most five tasks in
any of them. The name is stored exactly as printed here.

\subsection{Per-Family Gains}
\label{app:family-gains}

Round 2 against the round-0 baseline, hard, paired task by task. The
corresponding table for the terminal benchmark is in \app{app:terminal-map}.

\begin{apptabsm}
\begin{tabular}{@{} p{0.30\columnwidth} r p{0.30\columnwidth} c r@{}}
\toprule
\textbf{Family} & $n$ & \textbf{Composition}
 & \hdr{base}{$\to$\,r2} & \textbf{fixed} \\
\midrule
\idt{textworld\_receptacle-\ab based\_object} & 7
 & look\_at $\times$7 & 4\,$\to$\,7 & \gainval{3} \\
\idt{locate\_source\_receptacle} & 5
 & pick\_and\_place $\times$5 & 5\,$\to$\,5 & \naval \\
\idt{household\_object\_pick-\ab cool-\ab place} & 5
 & pick\_cool $\times$5 & 4\,$\to$\,5 & \gainval{1} \\
\idt{household\_object\_pick-\ab clean-\ab place} & 4
 & pick\_clean $\times$4 & 3\,$\to$\,4 & \gainval{1} \\
\idt{multi-\ab instance\_pick-\ab and-\ab place} & 3
 & pick\_two $\times$3 & 3\,$\to$\,3 & \naval \\
\idt{carryable\_household\_object} & 3
 & pick\_heat $\times$3 & 1\,$\to$\,3 & \gainval{2} \\
$\bigstar$ \idt{household\_pick-\ab and-\ab place:\ab \_locate} & 2
 & pick\_and\_place $\times$1 $+$ pick\_two $\times$1 & 2\,$\to$\,2 & \naval \\
$\bigstar$ \idt{household\_object\_pick-\ab cool-\ab place} & 2
 & pick\_cool $\times$1 $+$ pick\_heat $\times$1 & 2\,$\to$\,2 & \naval \\
\idt{household\_food\_object} & 2
 & pick\_heat $\times$2 & 2\,$\to$\,2 & \naval \\
\idt{bowl\_multi-\ab instance\_pick-\ab and-\ab place} & 2
 & pick\_two $\times$2 & 2\,$\to$\,2 & \naval \\
five singleton families & 1$\times$5
 & pick\_and\_place / pick\_clean $\times$2 / pick\_two / pick\_heat
 & 3\,$\to$\,4 & \gainval{1} \\
\bottomrule
\end{tabular}
\end{apptabsm}
\apptabcap{ALFWorld $\times$ MiniMax-M3. \emph{Composition} gives the dataset's
ground-truth types; $\bigstar$ marks a family that cuts across two of them.
Family names are generated by the compression stage and are not guaranteed
distinct: the two rows named \idt{household\_object\_pick-\ab cool-\ab place}
are different families with disjoint members.}{tab:o-alfworld}

\begin{apptabsm}
\begin{tabular}{@{}p{0.44\columnwidth} r c r@{}}
\toprule
\textbf{Family} & $n$ & \textbf{base\,$\to$\,r2} & \textbf{fixed} \\
\midrule
\idt{django\_numeric\_expression}      & 4 & 1\,$\to$\,2 & \gainval{1} \\
\idt{astropy\_ascii\_html}             & 2 & 2\,$\to$\,2 & \naval \\
\idt{django\_queryset\_filtering}      & 2 & 1\,$\to$\,2 & \gainval{1} \\
\idt{matplotlib\_image\_normalization} & 2 & 1\,$\to$\,2 & \gainval{1} \\
\idt{astropy\_fits\_diff}              & 1 & 1\,$\to$\,1 & \naval \\
\bottomrule
\end{tabular}
\end{apptabsm}
\apptabcap{SWE $\times$ GPT-5.4-mini; 9 tasks entered no cluster.}{tab:o-swe}

\begin{apptabsm}
\begin{tabular}{@{}p{0.44\columnwidth} r c r@{}}
\toprule
\textbf{Family} & $n$ & \textbf{base\,$\to$\,r2} & \textbf{fixed} \\
\midrule
\idt{exponential\_o-\ab minimal\_theory}    & 5 & 4\,$\to$\,5 & \gainval{1} \\
\idt{primitive\_permutation\_group}     & 4 & 1\,$\to$\,4 & \gainval{3} \\
\idt{asymptotically\_flat\_nonnegative} & 3 & 2\,$\to$\,3 & \gainval{1} \\
\idt{wythoff\_combinatorial\_game}      & 1 & 0\,$\to$\,1 & \gainval{1} \\
\idt{hankel-\ab matrix\_special-\ab fiber-\ab ring} & 1 & 1\,$\to$\,1 & \naval \\
\bottomrule
\end{tabular}
\end{apptabsm}
\apptabcap{LMB $\times$ MiniMax-M3; 39 tasks entered no cluster.}{tab:o-lmb}

The three per-benchmark tables are Table~\ref{tab:o-alfworld},
Table~\ref{tab:o-swe} and Table~\ref{tab:o-lmb}. Two notes on the read-out. More than one candidate prior is kept per family
while candidates are being generated, but exactly one per family is committed
to the library, so the prior count is the number of families plus one base
prior, and the counts above are the ones actually committed. Agreement with the
dataset's labels can only be measured on ALFWorld, the only one of the four
benchmarks carrying independent solution-type labels.

\section{Other Ways to Organise a Library}
\label{app:alternatives}

\subsection{Prior Repair}
\label{app:repair}

The second candidate route of \bsec{sec:commit} is append-only repair. Its net
effect, taken as the soft difference between two consecutive
sub-round evaluations of the same run, is \gainval{3.1} for DeepSeek,
\dropval{3.4} for GPT and \gainval{4.8} for MiniMax-M3. The three are of
comparable magnitude and the sign flips with the model, so repair is not
guaranteed to help and is adjudicated by the commit gate in real
execution. This is also why the three candidate routes are all sent into the
same gate rather than being given a fixed priority order.

\subsection{A Flat Retrieval Library}
\label{app:flat}

The other opposing form is a flat library that stores every historical local
skill and retrieves across tasks. That control was run on LMB's 15-task
validation split with the same model, the same injection channel and the same
retriever, changing only how the library is built.

\begin{apptabsm}
\begin{tabular}{@{}p{0.44\columnwidth} r c r@{}}
\toprule
\textbf{Library construction} & \hdr{en-}{tries}
 & \hdr{has}{$G_{\text{base}}$} & \hdr{val15}{soft} \\
\midrule
No-Skill                                      & \phantom{0}0 & \naval & 0.267 \\
Flat library, all local skills                & 53 & no  & 0.333 \\
Flat library, rebuilt after fixing retrieval  & 49 & yes & 0.333 \\
Flat library, construction three              & 54 & yes & 0.333 \\
Flat library, construction four               & 54 & yes & 0.467 \\
\textbf{Family library}, faithful LLM merge   & \textbf{22} & yes & 0.400 \\
\textbf{Family library}, another construction & \textbf{22} & yes & \textbf{0.467} \\
\bottomrule
\end{tabular}
\end{apptabsm}
\apptabcap{Flat versus family libraries on LMB val15.}{tab:p-flat}

The comparison is Table~\ref{tab:p-flat}. The family library reaches this
group's top score, 0.467, with $0.42\times$ the
entries, matching a flat library 2.4 times its size, and both family arms are
above the median of the four flat arms. This is the same kind of
statement as the compression ratio of \app{app:library-size}: the library is
smaller without being worse. The control is a 15-task validation split at a
single trial. The flat pool with retrieval in \btab{tab:ablation} is carried by the AWM column,
which is a read-out over all twelve cells.

%% file: references.bib
@inproceedings{jimenez2024swebench,
  author    = {Jimenez, Carlos E. and Yang, John and Wettig, Alexander and Yao, Shunyu and Pei, Kexin and Press, Ofir and Narasimhan, Karthik},
  title     = {{SWE}-bench: Can Language Models Resolve Real-World {GitHub} Issues?},
  booktitle = {International Conference on Learning Representations (ICLR)},
  year      = {2024}
}

@inproceedings{merrill2026terminalbench,
  author    = {Merrill, Mike A. and Shaw, Alexander G. and Carlini, Nicholas and Li, Boxuan and Raj, Harsh and Bercovich, Ivan and Shi, Lin and Shin, Jeong Yeon and Walshe, Thomas and Buchanan, E. Kelly and Shen, Junhong and Ye, Guanghao and Lin, Haowei and Poulos, Jason and Wang, Maoyu and others},
  title     = {Terminal-Bench: Benchmarking Agents on Hard, Realistic Tasks in Command Line Interfaces},
  booktitle = {International Conference on Learning Representations (ICLR)},
  year      = {2026},
  note      = {arXiv:2601.11868}
}

@misc{tbpro,
  author        = {Wang, Weixun and Xu, XiaoXiao and An, Wanhe and Dai, Fangwen and Gao, Wei and others},
  title         = {Let It Flow: Agentic Crafting on Rock and Roll, Building the {ROME} Model within an Open Agentic Learning Ecosystem},
  year          = {2025},
  eprint        = {2512.24873},
  archivePrefix = {arXiv}
}

@misc{lmb,
  author       = {He, Linyang and Yu, Qiyao and Dong, Hanze and Liao, Baohao and Xu, Xinxing and Goldblum, Micah and Bian, Jiang and Mesgarani, Nima},
  title        = {{LiveMathematicianBench}: A Live Benchmark for Mathematician-Level Reasoning with Proof Sketches},
  year         = {2026},
  eprint       = {2604.01754},
  archivePrefix = {arXiv}
}

@inproceedings{shridhar2021alfworld,
  author    = {Shridhar, Mohit and Yuan, Xingdi and C{\^o}t{\'e}, Marc-Alexandre and Bisk, Yonatan and Trischler, Adam and Hausknecht, Matthew},
  title     = {{ALFWorld}: Aligning Text and Embodied Environments for Interactive Learning},
  booktitle = {International Conference on Learning Representations (ICLR)},
  year      = {2021}
}

@misc{skillsbench,
  author        = {Li, Xiangyi and Chen, Wenbo and Liu, Yimin and Zheng, Shenghan and Chen, Xiaokun and others},
  title         = {{SkillsBench}: Benchmarking How Well Agent Skills Work Across Diverse Tasks},
  year          = {2026},
  eprint        = {2602.12670v1},
  archivePrefix = {arXiv}
}

@misc{agentskillssurvey,
  author       = {Zhou, Yingli and Wang, Shu and Su, Yaodong and Du, Wenchuan and Fang, Yixiang and Lin, Xuemin},
  title        = {A Comprehensive Survey on Agent Skills: Taxonomy, Techniques, and Applications},
  year         = {2026},
  eprint       = {2605.07358},
  archivePrefix = {arXiv}
}

@misc{autoskill,
  author       = {Yang, Yutao and Li, Junsong and Pan, Qianjun and Zhan, Bihao and Cai, Yuxuan and Du, Lin and Zhou, Jie and Chen, Kai and Chen, Qin and Li, Xin and Zhang, Bo and He, Liang},
  title        = {{AutoSkill}: Experience-Driven Lifelong Learning via Skill Self-Evolution},
  year         = {2026},
  eprint       = {2603.01145},
  archivePrefix = {arXiv}
}

@misc{trace2skill,
  author        = {Ni, Jingwei and Liu, Yihao and Liu, Xinpeng and Sun, Yutao and Zhou, Mengyu and Cheng, Pengyu and Wang, Dexin and Zhao, Erchao and Jiang, Xiaoxi and Jiang, Guanjun},
  title         = {{Trace2Skill}: Distill Trajectory-Local Lessons into Transferable Agent Skills},
  year          = {2026},
  eprint        = {2603.25158},
  archivePrefix = {arXiv}
}

@misc{skillopt,
  author       = {Yang, Yifan and Gong, Ziyang and Huang, Weiquan and Yang, Qihao and Zhou, Ziwei and Huang, Zisu and Li, Yan and Gao, Xuemei and Dai, Qi and Liu, Bei and Qiu, Kai and Yang, Yuqing and Chen, Dongdong and Yang, Xue and Luo, Chong},
  title        = {{SkillOpt}: Executive Strategy for Self-Evolving Agent Skills},
  year         = {2026},
  eprint       = {2605.23904},
  archivePrefix = {arXiv}
}

@inproceedings{acon,
  author    = {Kang, Minki and Chen, Wei-Ning and Han, Dongge and Inan, Huseyin A. and Wutschitz, Lukas and Chen, Yanzhi and Sim, Robert and Rajmohan, Saravan},
  title     = {{ACON}: Optimizing Context Compression for Long-horizon {LLM} Agents},
  booktitle = {International Conference on Machine Learning},
  year      = {2026}
}

@misc{memento,
  author       = {Zhou, Huichi and Chen, Yihang and Guo, Siyuan and Yan, Xue and Lee, Kin Hei and Wang, Zihan and Lee, Ka Yiu and Zhang, Guchun and Shao, Kun and Yang, Linyi and Wang, Jun},
  title        = {Memento: Fine-tuning {LLM} Agents without Fine-tuning {LLMs}},
  year         = {2025},
  eprint       = {2508.16153},
  archivePrefix = {arXiv}
}

@misc{mementoskills,
  author        = {Zhou, Huichi and Guo, Siyuan and Liu, Anjie and Yu, Zhongwei and Gong, Ziqin and Zhao, Bowen and Chen, Zhixun and Zhang, Menglong and Chen, Yihang and Li, Jinsong and Yang, Runyu and Liu, Qiangbin and Yu, Xinlei and Zhou, Jianmin and Wang, Na and Sun, Chunyang and Wang, Jun},
  title         = {Memento-Skills: Let Agents Design Agents},
  year          = {2026},
  eprint        = {2603.18743},
  archivePrefix = {arXiv}
}

@inproceedings{reme,
  author    = {Cao, Zouying and Deng, Jiaji and Yu, Li and Zhou, Weikang and Liu, Zhaoyang and Ding, Bolin and Zhao, Hai},
  title     = {Remember Me, Refine Me: A Dynamic Procedural Memory Framework for Experience-Driven Agent Evolution},
  booktitle = {Findings of the Association for Computational Linguistics: ACL 2026},
  pages     = {16803--16822},
  year      = {2026},
  doi       = {10.18653/v1/2026.findings-acl.829}
}

@misc{memorytransfer,
  author       = {Kim, Kangsan and Kang, Minki and Kim, Taeil and Yang, Yanlai and Ren, Mengye and Hwang, Sung Ju},
  title        = {Memory Transfer Learning: How Memories are Transferred Across Domains in Coding Agents},
  year         = {2026},
  eprint       = {2604.14004},
  archivePrefix = {arXiv}
}

@inproceedings{xskill,
  author    = {Jiang, Guanyu and Su, Zhaochen and Qu, Xiaoye and Fung, Yi R.},
  title     = {{XSkill}: Continual Learning from Experience and Skills in Multimodal Agents},
  booktitle = {International Conference on Machine Learning},
  year      = {2026}
}

@misc{skillos,
  author       = {Ouyang, Siru and Yan, Jun and Chen, Yanfei and Han, Rujun and Wang, Zifeng and Dalvi Mishra, Bhavana and Meng, Rui and Li, Chun-Liang and Jiao, Yizhu and Zha, Kaiwen and Shen, Maohao and Tirumalashetty, Vishy and Lee, George and Han, Jiawei and Pfister, Tomas and Lee, Chen-Yu},
  title        = {{SkillOS}: Learning Skill Curation for Self-Evolving Agents},
  year         = {2026},
  eprint       = {2605.06614},
  archivePrefix = {arXiv}
}

@misc{skillgraph,
  author       = {Li, Xiaoyuan and Li, Moxin and Bao, Keqin and Ma, Yubo and Wang, Wenjie and Liu, Dayiheng and Feng, Fuli},
  title        = {{SkillGraph}: Skill-Augmented Reinforcement Learning for Agents via Evolving Skill Graphs},
  year         = {2026},
  eprint       = {2605.12039},
  archivePrefix = {arXiv}
}

@misc{skillmas,
  author       = {Pan, Shuai and Liu, Yixiang and Gao, Jiaye and Gao, Te and Liu, Weiwen and Lin, Jianghao and Fu, Zhihui and Wang, Jun and Zhang, Weinan and Yu, Yong},
  title        = {{SkillMAS}: Skill Co-Evolution with {LLM}-based Multi-Agent System},
  year         = {2026},
  eprint       = {2605.09341},
  archivePrefix = {arXiv}
}

@misc{museautoskill,
  author       = {Lin, Huawei and Li, Peng and Song, Jie and Jiang, Fuxin and Zhang, Tieying},
  title        = {{MUSE-Autoskill}: Self-Evolving Agents via Skill Creation, Memory, Management, and Evaluation},
  year         = {2026},
  eprint       = {2605.27366},
  archivePrefix = {arXiv}
}

@inproceedings{kneedle,
  author    = {Satop{\"a}{\"a}, Ville and Albrecht, Jeannie and Irwin, David and Raghavan, Barath},
  title     = {Finding a ``Kneedle'' in a Haystack: Detecting Knee Points in System Behavior},
  booktitle = {31st International Conference on Distributed Computing Systems Workshops (ICDCSW)},
  pages     = {166--171},
  year      = {2011},
  doi       = {10.1109/ICDCSW.2011.20}
}

@misc{voyager,
  author        = {Wang, Guanzhi and Xie, Yuqi and Jiang, Yunfan and Mandlekar, Ajay and Xiao, Chaowei and Zhu, Yuke and Fan, Linxi and Anandkumar, Anima},
  title         = {Voyager: An Open-Ended Embodied Agent with Large Language Models},
  year          = {2023},
  eprint        = {2305.16291},
  archivePrefix = {arXiv}
}

@inproceedings{react,
  author    = {Yao, Shunyu and Zhao, Jeffrey and Yu, Dian and Du, Nan and Shafran, Izhak and Narasimhan, Karthik and Cao, Yuan},
  title     = {{ReAct}: Synergizing Reasoning and Acting in Language Models},
  booktitle = {International Conference on Learning Representations (ICLR)},
  year      = {2023}
}

@inproceedings{reflexion,
  author    = {Shinn, Noah and Cassano, Federico and Gopinath, Ashwin and Narasimhan, Karthik and Yao, Shunyu},
  title     = {Reflexion: Language Agents with Verbal Reinforcement Learning},
  booktitle = {Advances in Neural Information Processing Systems (NeurIPS)},
  volume    = {36},
  pages     = {8634--8652},
  year      = {2023}
}

@inproceedings{toolformer,
  author    = {Schick, Timo and Dwivedi-Yu, Jane and Dess{\`i}, Roberto and Raileanu, Roberta and Lomeli, Maria and Hambro, Eric and Zettlemoyer, Luke and Cancedda, Nicola and Scialom, Thomas},
  title     = {Toolformer: Language Models Can Teach Themselves to Use Tools},
  booktitle = {Advances in Neural Information Processing Systems},
  volume    = {36},
  pages     = {68539--68551},
  year      = {2023}
}

@inproceedings{sbert,
  author    = {Reimers, Nils and Gurevych, Iryna},
  title     = {{Sentence-BERT}: Sentence Embeddings Using Siamese {BERT}-Networks},
  booktitle = {Conference on Empirical Methods in Natural Language Processing (EMNLP-IJCNLP)},
  pages     = {3982--3992},
  year      = {2019},
  doi       = {10.18653/v1/D19-1410}
}

@inproceedings{reasoningbank,
  author    = {Ouyang, Siru and Yan, Jun and Hsu, I-Hung and Chen, Yanfei and Jiang, Ke and Wang, Zifeng and Han, Rujun and Le, Long T. and Daruki, Samira and Tang, Xiangru and Tirumalashetty, Vishy and Lee, George and Rofouei, Mahsan and Lin, Hangfei and Han, Jiawei and Lee, Chen-Yu and Pfister, Tomas},
  title     = {{ReasoningBank}: Scaling Agent Self-Evolving with Reasoning Memory},
  booktitle = {International Conference on Learning Representations},
  year      = {2026}
}

@misc{agentkb,
  author        = {Tang, Xiangru and Qin, Tianrui and Peng, Tianhao and Zhou, Ziyang and Shao, Daniel and Du, Tingting and Wei, Xinming and Xia, Peng and Wu, Fang and Zhu, He and Zhang, Ge and Liu, Jiaheng and Wang, Xingyao and Hong, Sirui and Wu, Chenglin and Cheng, Hao and Wang, Chi and Zhou, Wangchunshu},
  title         = {{Agent KB}: Leveraging Cross-Domain Experience for Agentic Problem Solving},
  year          = {2025},
  eprint        = {2507.06229},
  archivePrefix = {arXiv},
  primaryClass  = {cs.CL}
}

@misc{bottomup,
  author        = {Du, Jiawei and Wu, Jinlong and Chen, Yuzheng and Hu, Yucheng and Li, Bing and Zhou, Joey Tianyi},
  title         = {Rethinking Agent Design: From Top-Down Workflows to Bottom-Up Skill Evolution},
  year          = {2025},
  eprint        = {2505.17673},
  archivePrefix = {arXiv}
}

@misc{skillweaver,
  author        = {Zheng, Boyuan and Fatemi, Michael Y. and Jin, Xiaolong and Wang, Zora Zhiruo and Gandhi, Apurva and Song, Yueqi and Gu, Yu and Srinivasa, Jayanth and Liu, Gaowen and Neubig, Graham and Su, Yu},
  title         = {{SkillWeaver}: Web Agents Can Self-Improve by Discovering and Honing Skills},
  year          = {2025},
  eprint        = {2504.07079},
  archivePrefix = {arXiv}
}

@misc{skillflow,
  author        = {Zhang, Ziao and Shi, Kou and Huang, Shiting and Nie, Avery and Zeng, Yu and Zhao, Yiming and Fang, Zhen and Su, Qishen and Qiu, Haibo and Yang, Wei and Ren, Qingnan and Zou, Shun and Huang, Wenxuan and Chen, Lin and Chen, Zehui and Zhao, Feng},
  title         = {{SkillFlow}: Benchmarking Lifelong Skill Discovery and Evolution for Autonomous Agents},
  year          = {2026},
  eprint        = {2604.17308},
  archivePrefix = {arXiv}
}

@article{textgrad,
  author  = {Yuksekgonul, Mert and Bianchi, Federico and Boen, Joseph and Liu, Sheng and Lu, Pan and Huang, Zhi and Guestrin, Carlos and Zou, James},
  title   = {Optimizing Generative {AI} by Backpropagating Language Model Feedback},
  journal = {Nature},
  volume  = {639},
  number  = {8055},
  pages   = {609--616},
  year    = {2025},
  doi     = {10.1038/s41586-025-08661-4}
}

@inproceedings{dspy,
  author    = {Khattab, Omar and Singhvi, Arnav and Maheshwari, Paridhi and Zhang, Zhiyuan and Santhanam, Keshav and Vardhamanan, Sri and Haq, Saiful and Sharma, Ashutosh and Joshi, Thomas T. and Moazam, Hanna and Miller, Heather and Zaharia, Matei and Potts, Christopher},
  title     = {{DSPy}: Compiling Declarative Language Model Calls into Self-Improving Pipelines},
  booktitle = {International Conference on Learning Representations},
  year      = {2024}
}

@inproceedings{opro,
  author    = {Yang, Chengrun and Wang, Xuezhi and Lu, Yifeng and Liu, Hanxiao and Le, Quoc V. and Zhou, Denny and Chen, Xinyun},
  title     = {Large Language Models as Optimizers},
  booktitle = {International Conference on Learning Representations (ICLR)},
  year      = {2024}
}

@inproceedings{selfrefine,
  author    = {Madaan, Aman and Tandon, Niket and Gupta, Prakhar and Hallinan, Skyler and Gao, Luyu and Wiegreffe, Sarah and Alon, Uri and Dziri, Nouha and Prabhumoye, Shrimai and Yang, Yiming and Gupta, Shashank and Majumder, Bodhisattwa Prasad and Hermann, Katherine and Welleck, Sean and Yazdanbakhsh, Amir and Clark, Peter},
  title     = {Self-Refine: Iterative Refinement with Self-Feedback},
  booktitle = {Advances in Neural Information Processing Systems (NeurIPS)},
  volume    = {36},
  pages     = {46534--46594},
  year      = {2023}
}

@inproceedings{expel,
  author    = {Zhao, Andrew and Huang, Daniel and Xu, Quentin and Lin, Matthieu and Liu, Yong-Jin and Huang, Gao},
  title     = {{ExpeL}: {LLM} Agents Are Experiential Learners},
  booktitle = {AAAI Conference on Artificial Intelligence (AAAI)},
  year      = {2024}
}

@inproceedings{sweagent,
  author    = {Yang, John and Jimenez, Carlos E. and Wettig, Alexander and Lieret, Kilian and Yao, Shunyu and Narasimhan, Karthik and Press, Ofir},
  title     = {{SWE}-agent: Agent-Computer Interfaces Enable Automated Software Engineering},
  booktitle = {Advances in Neural Information Processing Systems (NeurIPS)},
  volume    = {37},
  pages     = {50528--50652},
  year      = {2024}
}

@inproceedings{agentbench,
  author    = {Liu, Xiao and Yu, Hao and Zhang, Hanchen and Xu, Yifan and Lei, Xuanyu and Lai, Hanyu and Gu, Yu and Ding, Hangliang and Men, Kaiwen and Yang, Kejuan and Zhang, Shudan and Deng, Xiang and Zeng, Aohan and Du, Zhengxiao and Zhang, Chenhui and Shen, Sheng and Zhang, Tianjun and Su, Yu and Sun, Huan and Huang, Minlie and Dong, Yuxiao and Tang, Jie},
  title     = {{AgentBench}: Evaluating {LLMs} as Agents},
  booktitle = {International Conference on Learning Representations (ICLR)},
  year      = {2024}
}

@inproceedings{gaia,
  author    = {Mialon, Gr{\'e}goire and Fourrier, Cl{\'e}mentine and Swift, Craig and Wolf, Thomas and LeCun, Yann and Scialom, Thomas},
  title     = {{GAIA}: A Benchmark for General {AI} Assistants},
  booktitle = {International Conference on Learning Representations (ICLR)},
  year      = {2024}
}

@article{consensus,
  author  = {Monti, Stefano and Tamayo, Pablo and Mesirov, Jill and Golub, Todd},
  title   = {Consensus Clustering: A Resampling-Based Method for Class Discovery and Visualization of Gene Expression Microarray Data},
  journal = {Machine Learning},
  volume  = {52},
  pages   = {91--118},
  year    = {2003},
  doi     = {10.1023/A:1023949509487}
}

@article{silhouette,
  author  = {Rousseeuw, Peter J.},
  title   = {Silhouettes: A Graphical Aid to the Interpretation and Validation of Cluster Analysis},
  journal = {Journal of Computational and Applied Mathematics},
  volume  = {20},
  pages   = {53--65},
  year    = {1987},
  doi     = {10.1016/0377-0427(87)90125-7}
}

@article{ward,
  author = {Ward, Jr., Joe H.},
  title   = {Hierarchical Grouping to Optimize an Objective Function},
  journal = {Journal of the American Statistical Association},
  volume  = {58},
  number  = {301},
  pages   = {236--244},
  year    = {1963},
  doi     = {10.1080/01621459.1963.10500845}
}

@inproceedings{memp,
  author    = {Fang, Runnan and Liang, Yuan and Wang, Xiaobin and Wu, Jialong and Qiao, Shuofei and Xie, Pengjun and Huang, Fei and Chen, Huajun and Zhang, Ningyu},
  title     = {{Memp}: Exploring Agent Procedural Memory},
  booktitle = {Findings of the Association for Computational Linguistics: ACL 2026},
  pages     = {17490--17502},
  year      = {2026},
  doi       = {10.18653/v1/2026.findings-acl.866}
}

@inproceedings{awm,
  author    = {Wang, Zora Zhiruo and Mao, Jiayuan and Fried, Daniel and Neubig, Graham},
  title     = {Agent Workflow Memory},
  booktitle = {Proceedings of the 42nd International Conference on Machine Learning},
  volume    = {267},
  pages     = {63897--63911},
  year      = {2025}
}

@misc{dyncheatsheet,
  author        = {Suzgun, Mirac and Yuksekgonul, Mert and Bianchi, Federico and Jurafsky, Dan and Zou, James},
  title         = {Dynamic Cheatsheet: Test-Time Learning with Adaptive Memory},
  year          = {2025},
  eprint        = {2504.07952},
  archivePrefix = {arXiv}
}

@misc{swebskills,
  author        = {Han, Tingxu and Zhang, Yi and Song, Wei and Fang, Chunrong and Chen, Zhenyu and Sun, Youcheng and Hu, Lijie},
  title         = {{SWE-Skills-Bench}: Do Agent Skills Actually Help in Real-World Software Engineering?},
  year          = {2026},
  eprint        = {2603.15401},
  archivePrefix = {arXiv}
}

@misc{qwen3embed,
  author        = {Zhang, Yanzhao and Li, Mingxin and Long, Dingkun and Zhang, Xin and Lin, Huan and Yang, Baosong and Xie, Pengjun and Yang, An and Liu, Dayiheng and Lin, Junyang and Huang, Fei and Zhou, Jingren},
  title         = {{Qwen3} Embedding: Advancing Text Embedding and Reranking Through Foundation Models},
  year          = {2025},
  eprint        = {2506.05176},
  archivePrefix = {arXiv}
}
